\documentclass{article}

\usepackage[margin=1.2in]{geometry}

\usepackage{natbib}
\usepackage{authblk} % author affiliations

\usepackage{amsmath}
\usepackage{amssymb}
\usepackage{amsthm}
\usepackage{mathtools}

\DeclareMathOperator*{\argmin}{arg\,min}

\usepackage{microtype}
\usepackage{graphicx}
\usepackage{booktabs} % tables
\usepackage{xcolor}
\usepackage{soul} % highlight
\usepackage{subcaption}
\usepackage{float}

\definecolor{cornflowerblue}{rgb}{0.39, 0.58, 0.93}
\definecolor{interaction_color}{HTML}{FF8F00}
\definecolor{oellm_blue}{HTML}{0192E0}
\definecolor{oellm_dark_blue}{HTML}{003EA8}
\definecolor{cool_pink}{HTML}{cc87e0}
\definecolor{oellm_orange}{HTML}{DB7401}

\usepackage{hyperref}
\hypersetup{
    colorlinks=true,
    linkcolor=oellm_blue,
    filecolor=magenta,      
    urlcolor=cool_pink,
    citecolor=oellm_blue,%teal,
    pdftitle={OELLM Scaling},
    pdfpagemode=FullScreen,
}

\usepackage{fontawesome}
\usepackage{makecell}

\makeatletter
\renewcommand\AB@affilsepx{, \protect\Affilfont}
\makeatother
\renewcommand\Affilfont{\small} % or \small 

\title{\includegraphics[width=0.15\textwidth]{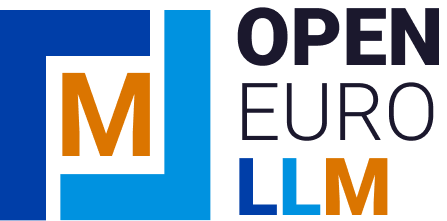}\\[1em]Deriving Scaling Laws for OpenEuroLLM Models: \\Learning Rate, Batch Size and Loss}

\author[1,9]{Niccolò Ajroldi}
\author[2,9]{Diana Alexandra Onuțu}
\author[1,3,9]{Haider Al-Tahan}
\author[1,4,6,7,9]{Jörg Franke}
\author[8,9]{Sampo Pyysalo}
\author[5,6,7,9]{Jenia Jitsev}
\author[1,9]{Aaron Klein}

\affil[1]{ELLIS Institute Tübingen}
\affil[2]{Eindhoven University of Technology}
\affil[3]{Georgia Institute of Technology}
\affil[4]{University of Freiburg}
\affil[5]{Jülich Supercomputing Center (JSC)}
\affil[6]{LAION}
\affil[7]{Open-$\Psi$ (Open-Sci) Collective}
\affil[8]{University of Turku}
\affil[9]{OpenEuroLLM}

\begin{document}

\date{}
\maketitle

\begingroup
\renewcommand{\thefootnote}{}
\footnotetext{%
    \centering
    \raisebox{-0.3em}{\includegraphics[height=1.1em]{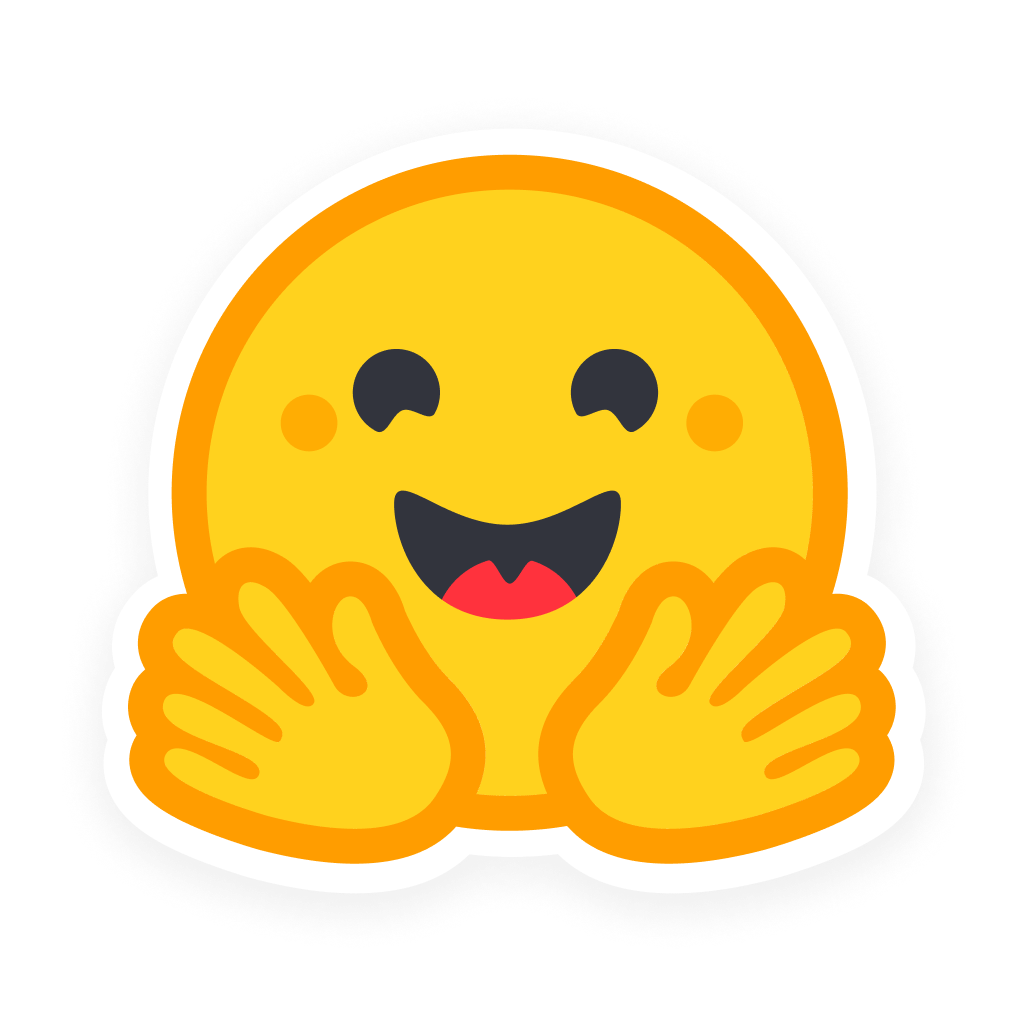}}~
    \href{https://huggingface.co/openeurollm/dense_english_scaling_laws/}
    {\texttt{openeurollm/dense\_english\_scaling\_laws}}
    \qquad
    \faGithub~
    \href{https://github.com/OpenEuroLLM/dense_english_scaling_laws}
    {\texttt{OpenEuroLLM/dense\_english\_scaling\_laws}}
}
\endgroup

\begin{abstract}
We study the scaling behavior of learning rate and batch size in pretraining dense large language models on English-prevalent corpora. Beyond scaling \textit{jointly optimal} learning rates and batch sizes, we investigate their \textit{marginal} evolution with model capacity and data scale and develop a model that captures these relationships. As we employ a Warmup-Stable-Decay learning rate schedule, we further investigate the gains from learning rate annealing over a broad range of hyperparameters settings, models and data budgets, and whether the optimal learning rate and batch size \textit{transfer} between the stable and decay phases. Finally, we characterize the dependence of loss on model capacity and dataset size, evaluating recently proposed scaling forms that explicitly model their interaction. We find these approaches particularly effective at capturing both undertraining and overtraining regimes across our experiments. This study establishes a first baseline and scaling procedure for the development of future OpenEuroLLM models. We open-source the complete collection of pretraining runs used in this study.
\end{abstract}

%%%%%%%%%%%%%%%%%%%%%%%%%%%%%%%%%%%%%%%%%%%%%%
%%%%%%%%%%%%%%%%%%%%%%%%%%%%%%%%%%%%%%%%%%%%%%

\section{Introduction}
\label{sec:intro}

\vspace{-2mm}\paragraph{Scaling optimal hyperparameters.}
Pretraining Large Language Models (LLMs) at scale requires making informed decisions about how models, data, and training configurations should grow.  Among these, a critical component of successful pretraining recipes is the choice of learning rate and batch size, which not only affects training efficiency but is also crucial for deriving accurate scaling laws~\citep{lourie2026smallscaleexperimentsyet}. 
To ensure that measured performance reflects the architecture and dataset scaling properties rather than hyperparameter mistuning, it is important to understand how optimal learning rates and batch sizes themselves scale with models size and training horizons.
Whereas early work focused on the \textit{critical} batch size~\citep{mccandlish2018empiricalmodellargebatchtraining, merrill2025critical, bergsma2025powerlinesscalinglaws}, more recent approaches have investigated how the \textit{optimal} batch size scales with number of parameters, data, compute, or combinations thereof \citep{hu2024minicpm, deepseekv12024, li2025steplaw, rutte2026scaling_diffusion}.
On the other hand, learning rate (LR) scaling has been studied both theoretically and empirically, with muP~\citep{yang_2021_muP} providing a principled framework for transferring optimal learning rates across model scales. In parallel, several works have characterized scaling laws for the optimal learning rate, both with and without muP parameterization~\citep{shen2024powerschedulerbatchsize, porian2025resolving}.
In this report, we present an extensive study of learning rate and batch size scaling. Rather than considering only their joint optimum, we distinguish between their \textit{joint} and \textit{marginal} scaling behavior, and develop models for both. In addition, leveraging this extensive set of training runs, we investigate how optimal hyperparameters are systematically affected by learning rate \textit{annealing} under modern Warmup-Stable-Decay (WSD) learning rate schedule.

\vspace{-2mm}\paragraph{Performance prediction.}
With appropriate hyperparameters in place, scaling laws can characterize how model performance evolves with scale. Laws of this kind have been used to predict performance at larger scales, to derive compute-optimal boundaries~\citep{Kaplan2020ScalingLF, hoffmann2022training}, and as a mean of comparison to evaluate models~\citep{nezhurina2025scaling} and datasets~\citep{nezhurina2025opensciref001openreproduciblereference}.
To better capture the dependencies between model and data scale, and better account for overtrained and undertrained models, recent works have explored functional forms beyond the well-known Chinchilla formulation~\citep{hoffmann2022training,videau2026skaling}. In this work, we explore this dependence by fitting to a suite of training runs spanning a broad range of model and data scales. As developing scaling laws is challenging~\citep{li2025misfittingsurveyscalinglaws} and their results can be difficult to reproduce~\citep{besiroglu2024chinchilla}, we open-source our training data and code to foster further research.

% Developing scaling laws that extrapolate reliably to larger scales is challenging~\citep{li2025misfittingsurveyscalinglaws}. 

\bigskip
The remainder of this technical report is organized as follows. \autoref{sec:experimental_setting} presents the experimental setting; \autoref{sec:scaling_hps} derives scaling laws for optimal learning rate and batch size; \autoref{sec:annealing_effect} analyzes the effect of LR annealing on loss and hyperparameters; \autoref{sec:scaling_loss} studies scaling laws for cross entropy loss; \autoref{sec:downstream_evals} evaluates models on downstream tasks; and \autoref{sec:conclusions} concludes.

%%%%%%%%%%%%%%%%%%%%%%%%%%%%%%%%%%%%%%%%%%%%%%

% \section{Design of Experiments}
% \subsection{Experimental Design}
\section{Experimental Setting}
\label{sec:experimental_setting}

\vspace{-2mm}\paragraph{Data.}
We train all models on the high-quality subset of the \texttt{Nemotron-CC} dataset~\citep{nemotron_cc_2025}, motivated by the strong empirical performance of such data mix reported by \citet{nezhurina2025opensciref001openreproduciblereference}.
Our experiments span up to $300$ billion tokens (BT), and use the \texttt{GPT-NeoX-20B} tokenizer~\citep{black_etal_2022_gptneox}. 
Unless otherwise specified, all experiments use the same data stream and initialization seed. Validation loss is measured on a common held-out set containing $204{,}800$ sequences ($0.838$BT). Throughout this work, “loss” and $L$ always refer to the \textit{validation loss}.

\vspace{-2mm}\paragraph{Models.}
We train dense decoder-only transformers ranging from 47 million (M) to 1.7 billion (B) parameters, including embedding parameters. All models use a vocabulary size of $50304$, FFN expansion factor $4$, GLU activations~\citep{shazeer2020gluvariantsimprovetransformer}, attention bias terms, QK normalization~\citep{henry2020querykeynormalizationtransformers}, and tied input-output embeddings, akin to the model designs of \citet{nezhurina2025opensciref001openreproduciblereference}. 
Model sizes are given in Table~\ref{tab:model_architectures} and spaced approximately geometrically in parameter count, with a fixed head dimension of 64 and a width-to-depth ratio that increases moderately with scale, following common practice for decoder-only transformers~\citep{Kaplan2020ScalingLF}.

\begin{table}[h]
\centering
\caption{Model architectures.}
\label{tab:model_architectures}
\begin{tabular}{lcccccc}
\toprule
% \midrule
Model dimension & 384 & 576 & 896 & 1280 & 1536 & 2048 \\
Attention heads & 6 & 9 & 14 & 20 & 24 & 32 \\
Layers & 12 & 18 & 20 & 20 & 24 & 24 \\
FFN dimension & 1536 & 2304 & 3584 & 5120 & 6144 & 8192 \\
\midrule
Total parameters & 47.6M & 124.5M & 301.9M & 588.5M & 983.1M & 1.713B \\
\bottomrule
\end{tabular}
\end{table}

\vspace{-2mm}\paragraph{Training configuration.}
Training is carried out with AdamW~\citep{loshchilov2019adamw}, coupled with a Warmup-Stable-Decay (WSD, or trapezoidal) learning rate schedule~\citep{zhai2022scalingvisiontransformers,hu2024minicpm}. We train for up to $300$BT and perform intermediate decays at $D=6,12,20,30,50,80,120,200$BT. We decay the learning rate over the final $20\%$ of each target token budget $D$, always annealing to a final learning rate of $10^{-5}$. We employ WSD due to its lower training cost~\citep{hagele2024scaling} compared to a cosine schedule, estimating a cost reduction of roughly \textit{half} for our experimental configuration. We use a fixed warmup of 2000 steps, following common practice~\citep{bakouch2025smollm3}. We train all models on AMD Instinct MI250X GPUs, using \texttt{Megatron-LM}~\citep{megatron-lm}, and use \texttt{autoexperiment}\footnote{\url{https://github.com/SLAMPAI/autoexperiment}} orchestrator to schedule experiments.
%The full collection of experiments amounts to roughly $\sim111.72$ZFLOPs\footnote{estimated as $C=6ND$}, out of which 78.6\% is required for annealing experiments.

\vspace{-2mm}\paragraph{Hyperparameter grid.}
We employ batch sizes $b$ $\log_2$-spaced in the range $2^4$ to $2^{10}$, and learning rates values $\eta \in \{.00025,.0005,.001,.002,.004\}$, forming a learning rate–batch size grid for each $N,D$ pair\footnote{We avoid training at all $(\eta,b)$ combinations for each $(N, D)$; instead, we prefer larger batch sizes for larger token budgets and smaller $b$ for smaller $D$ values, always ensuring that the empirical optimum ($\eta^\star, b^\star$) does not lie on the boundary of the grid.}. Since we train across different batch sizes $b$ with a fixed warmup duration, we exclude $(b,D)$ combinations that would result in 
%an LR schedule with 
a warmup longer than $30\%$ of $D$, to avoid degenerate LR schedules and ensure the peak learning rate is reached and sustained for a meaningful portion of training. 

\vspace{-2mm}\paragraph{Open-source.}
We release the complete collection of pretraining runs used in this study, including both model checkpoints\footnote{\url{https://huggingface.co/openeurollm/dense_english_scaling_laws}} and the corresponding loss and downstream evaluations \footnote{\url{https://github.com/OpenEuroLLM/dense_english_scaling_laws}}. The collection spans a broad range of model sizes, token budgets, learning rates, and batch sizes, providing a resource for reproducing our analyses and studying the scaling behavior of optimization hyperparameters.

%%%%%%%%%%%%%%%%%%%%%%%%%%%%%%%%%%%%%%%%%%%%%%%%%%%%%%%%%%%%%%%%%%%%%%%%%%%%%

\section{Scaling Hyperparameters}
\label{sec:scaling_hps}

We aim to study how the optimal batch size and learning rate vary across model sizes and token budgets. 
Before proceeding, we describe the loss smoothing procedure used to obtain more reliable estimates of the optimal hyperparameters.
With this procedure in place,
we consider \textit{both} the jointly optimal learning rate and batch size and the optima obtained when fixing one hyperparameter and optimizing the other.  We then characterize these trends by fitting power laws and evaluate their predictive accuracy on held-out 1.7B models trained across a range of data budgets. Unless otherwise specified, confidence and prediction intervals are computed analytically under the fitted parametric model using a significance level of 0.05.

\subsection{Smoothing}
\label{sec:smoothing}

\begin{figure}[h]
    \centering
    \makebox[\linewidth][c]{%
        \begin{subfigure}[c]{0.25\linewidth}
            \centering
            \includegraphics[width=\linewidth]{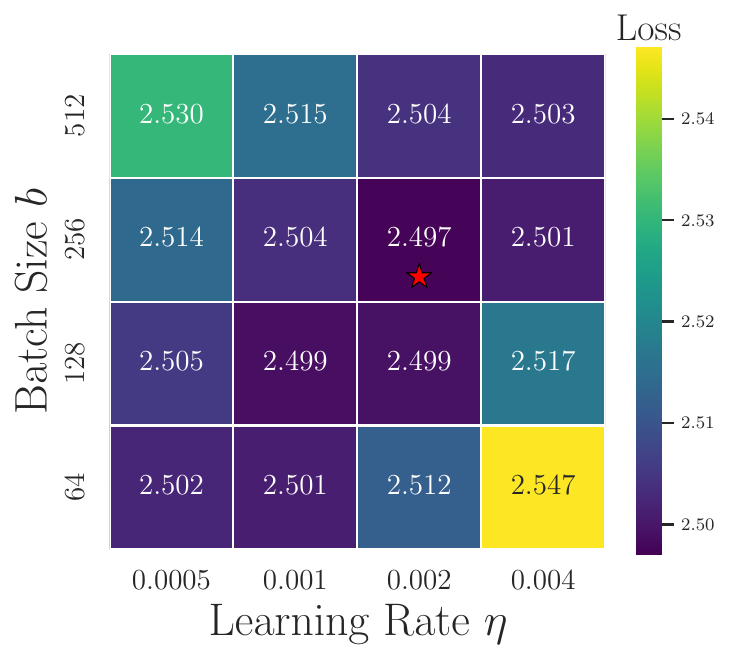}
        \end{subfigure}
        \hspace{0.01\linewidth}
        \begin{subfigure}[c]{0.25\linewidth}
            \centering
            \includegraphics[width=\linewidth]{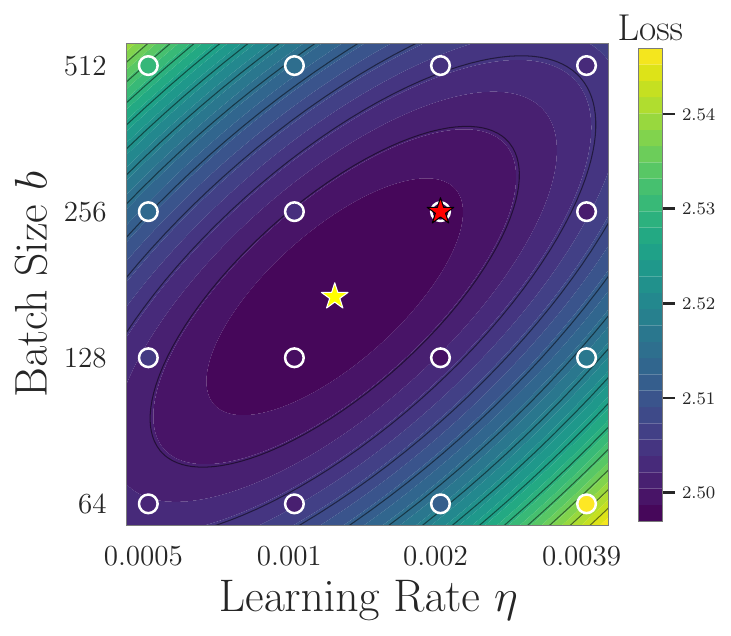}
        \end{subfigure}
        \hspace{0.01\linewidth}
        \begin{subfigure}[c]{0.335\linewidth}
            \centering
            \includegraphics[width=\linewidth]{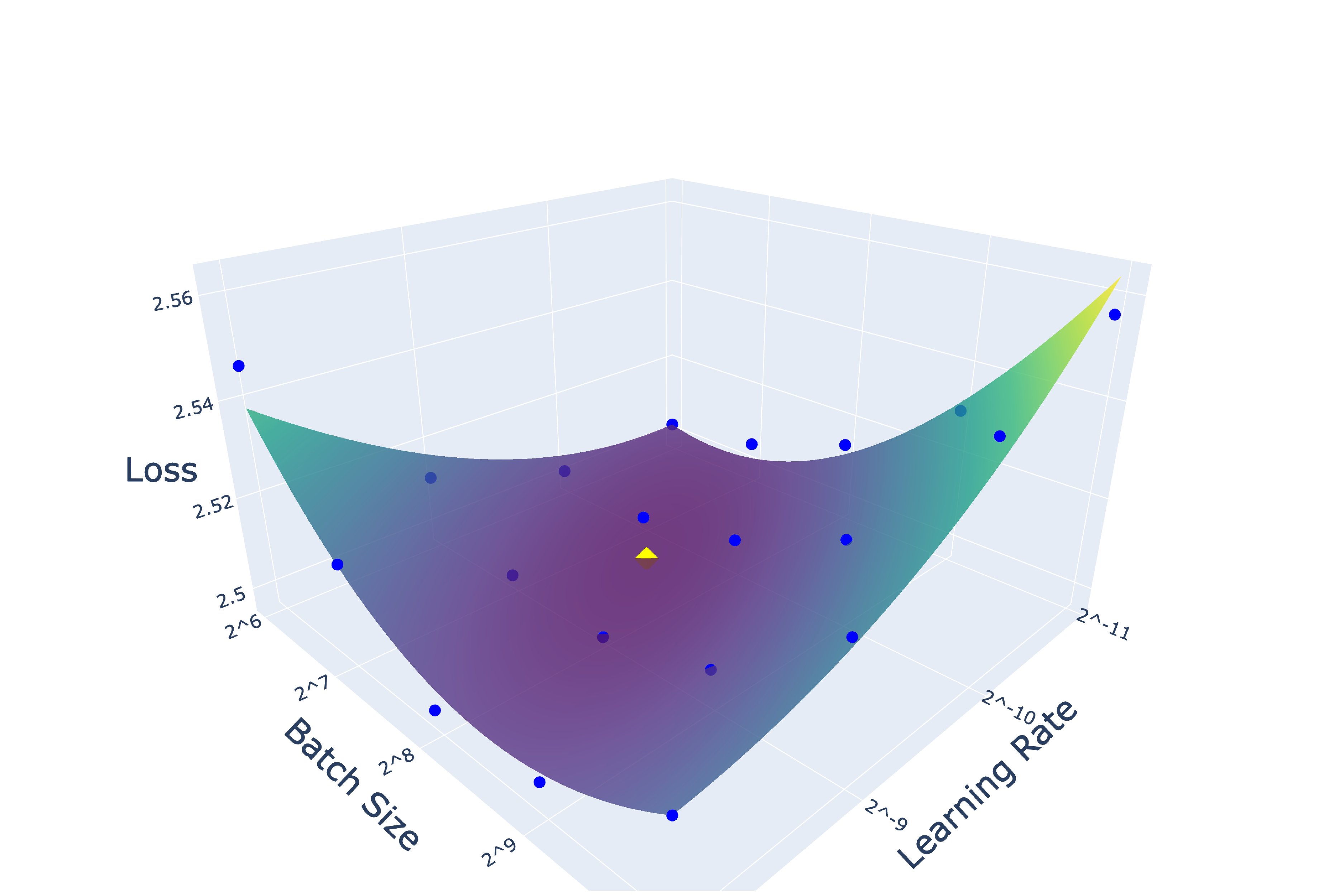}
        \end{subfigure}%
    }
    \caption{Validation loss vs learning rate and batch size for $N=0.13$B, $D=120$BT. \textbf{First column}: raw observations and empirical minima (\textcolor[HTML]{FF0200}{$\star$}). \textbf{Second and third columns}: estimated quadratic function in $\eta, b$ (\autoref{eq:quadratic_2d}), together with the predicted joint optima (\textcolor[HTML]{E1D611}{$\star$}) from \autoref{eq:joint_optima}.}
    \label{fig:smooth}
\end{figure}

To obtain reliable estimates of the optimal hyperparameters for each model size and data budget, we smooth the validation loss surface. This serves two purposes. 
First, as loss measurements are collected on a relatively coarse grid of learning rates and batch sizes, smoothing provides a continuous approximation of the loss surface, that can better capture shifts in the optimum across $N$ and $D$.
Second, as validation loss is inherently noisy due to data ordering, random initialization, and finite validation-set sampling, we argue that smoothing can reduce the variance of the optimal hyperparameter estimates relative to empirical optima selection. We refer to \autoref{sec:smoothing_robustness} for an analysis of the sensitivity and robustness of the smoothing procedure.

In line with previous work \citep{li2025steplaw, lourie2025hyperparameter, lourie2026smallscaleexperimentsyet}, we find that the loss can be well approximated by a quadratic function of the learning rate and batch size around the optima:
\begin{equation}
    L(\eta,b|N,D) = \beta_0 + \beta_1\eta + \beta_2\eta^2 + \beta_3 b + \beta_4 b^2 + \beta_5 \eta b + \epsilon, \qquad \forall N,D.
    \label{eq:quadratic_2d}
\end{equation}
We estimate $\beta_0,\ldots,\beta_5$ for each $N,D$ using Ordinary Least Squares (OLS), and use the fitted values to predict the optimal hyperparameters:
\begin{equation}
    \eta^\star, b^\star | N,D = \argmin_{\eta, b} \hat{L}(\eta,b|N,D), \qquad \forall N,D.
    \label{eq:joint_optima}
\end{equation}
Fixing one hyperparameter, $\eta$ or $b$, allows us to obtain a quadratic loss function in the other, which we can use to predict the optimal hyperparameters conditionally on one another\footnote{Notice that, instead of slicing \autoref{eq:quadratic_2d} to obtain one-dimensional quadratic fits, one could fit independent one-dimensional quadratics, as in \citet{bjorck2025scalingoptimallrtoken}. We instead use the two-dimensional fit because some $b$ values have only 3--4 learning rate observations, whereas each $N,D$ pair has at least 15 points in the $\eta,b$ grid, providing more degrees of freedom and thus more robust estimates. \autoref{fig:smooth} shows an example of the data and the fits.}:
\begin{align}
    \eta^\star | b,N,D = \argmin_{\eta} \hat{L} (\eta,b|N,D) \ \forall b,N,D; \qquad
    b^\star | \eta, N,D = \argmin_{b} \hat{L} (\eta,b|N,D) \ \forall \eta,N,D.
    \label{eq:individual_optima}
\end{align}

Note that smoothing introduces an additional source of uncertainty, since the smoothed estimates are themselves obtained through a fitting process. We acknowledge this limitation and, for simplicity, condition on the smoothed estimates when quantifying uncertainty in the subsequent scaling law fits.

% ------------------------------------------------------------------

\subsection{Scaling Jointly Optimal Learning Rate and Batch Size}
\label{sec:scaling_jointly_lr_bsz}

\begin{figure}
    \centering
    \includegraphics[width=.95\linewidth]{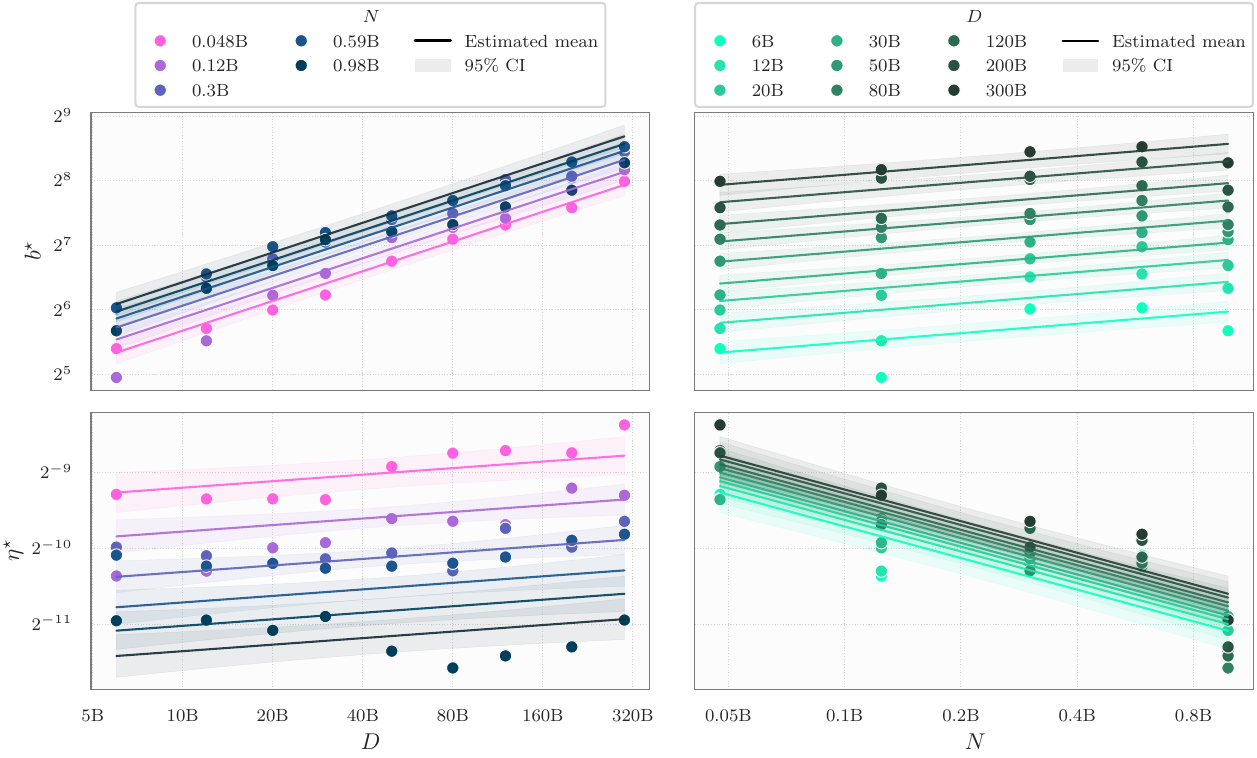}
    \caption{Scaling laws for jointly optimal batch size (\autoref{eq:joint_lr_bsz_scaliing_BSZ}) and learning rate (\autoref{eq:joint_lr_bsz_scaliing_LR}).}
    \label{fig:joint_hp_scaling}
\end{figure}

\begin{figure}
    \begin{subfigure}{\linewidth}
        \centering
        \includegraphics[width=.8\linewidth]{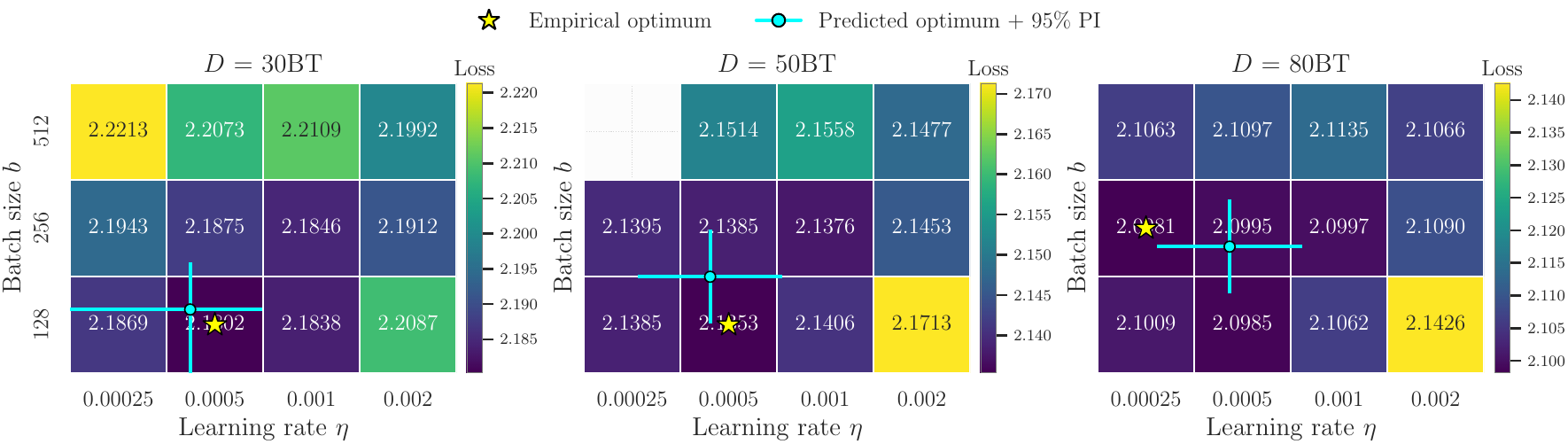}
        \caption{Empirical loss values.}
        \label{fig:joint_hp_scaling_preciction_accuracy_grid}
    \end{subfigure}
    \begin{subfigure}{\linewidth}
        \centering
        \includegraphics[width=.8\linewidth]{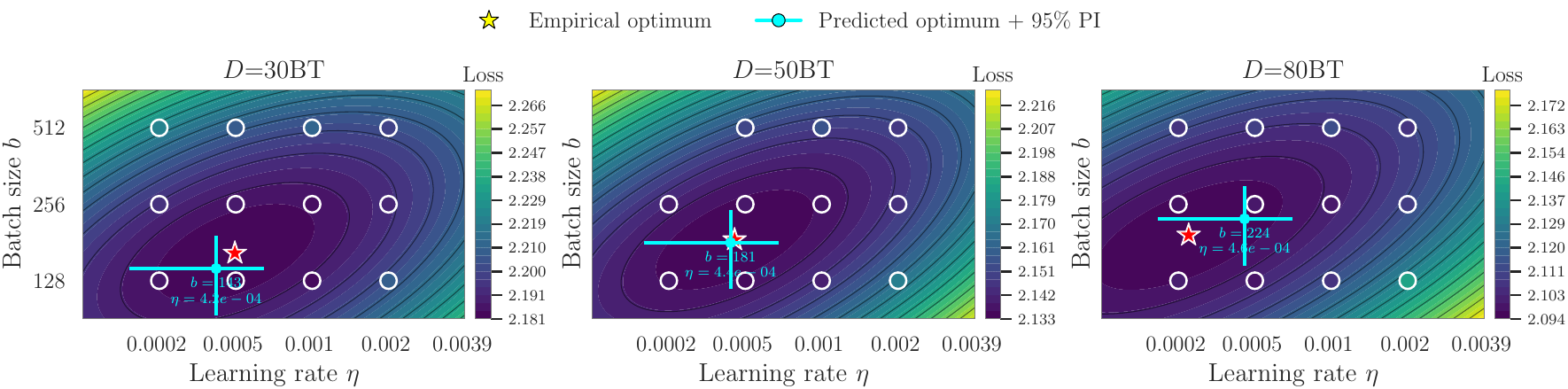}
        \caption{Quadratic approximation of the same loss surface.}
        \label{fig:joint_hp_scaling_preciction_accuracy_contour}
    \end{subfigure}
    \caption{\textbf{Prediction accuracy on held-out 1.7B model} of joint optimal learning rate and batch size scaling laws. We report empirical optima (\textcolor[HTML]{FF0200}{$\star$}) and minima of the fitted quadratic surface (\textcolor[HTML]{E1D611}{$\star$}). Cyan circles and error bars denote the predicted optima with marginal parametric $95\%$ prediction intervals. }
    \label{fig:joint_hp_scaling_preciction_accuracy}
\end{figure}

% Motivation, literature
% We first investigate how the jointly optimal learning rate and batch size scale with model size and training data. The goal is to characterize these trends to predict near-optimal hyperparameters for larger training runs without an expensive hyperparameter search. Such setting has been the primary focus of recent work on hyperparameter scaling laws, such as \citet{deepseekv12024, li2025steplaw, bergsma2025powerlinesscalinglaws, porian2025resolving}.

As previously discussed, and similarly to \citet{li2025steplaw}, we consider the jointly optimal learning rate and batch size, $(\eta^\star, b^\star)$, obtained from the quadratic smoothing in \autoref{eq:joint_optima}. We observe approximately linear trends in log-space and therefore fit the following models:
\begin{align}
    \log b^\star(N,D) &= \alpha_1 + \beta_1 \log N + \gamma_1 \log D + \varepsilon_1,
    \label{eq:joint_hp_scaling_bsz}\\
    \log \eta^\star(N,D) &= \alpha_2 + \beta_2 \log N + \gamma_2 \log D + \varepsilon_2 ,
    \label{eq:joint_hp_scaling_lr}
\end{align}
where $\varepsilon_1$ and $\varepsilon_2$ denote the regression errors. The two models are estimated independently and thus do not explicitly model any covariance between residuals. 

\vspace{-2mm}\paragraph{Batch size.}
The fitted model in \autoref{eq:joint_hp_scaling_bsz} explains $93.6\%$ of the variance in the observed optimal batch sizes. We find a strong positive scaling of the optimal batch size with respect to $D$ (0.46). Differently from \citep{li2025steplaw} and \citep{bergsma2025powerlinesscalinglaws}, we find a weak but positive and statistically significant effect of $N$ on the optimal batch size ($\hat{\beta_1} = 0.145$). We depict the data and fitted models in \autoref{fig:joint_hp_scaling}. Rewriting the fitted equations as power laws, we obtain the following equation:
\begin{equation}
    \hat{b}^\star | N,D = 0.000099\, N^{0.145} D^{0.460}.
    \label{eq:joint_lr_bsz_scaliing_BSZ}
\end{equation}

\vspace{-2mm}\paragraph{Learning rate.}
Despite some clear deviations from the fits in \autoref{fig:joint_hp_scaling}, the linear model captures 79.2\% of the variance in the data. The optimal learning rate scales strongly with respect to $N$ (0.417) and exhibits a weaker dependence on $D$ (0.0862), which nevertheless appears statistically significant after bootstrap testing, yielding a \textit{p-value} of 0.0055. Comparing the estimated coefficients with~\citet{li2025steplaw}, we find smaller scaling exponents for both $D$ and $N$. The fitted coefficients with respect to $N$ are however close to the work from~\citet{porian2025resolving}. The final proposed model is:
\begin{equation}
    \hat{\eta}^\star | N,D = 0.371396\, N^{-0.417} D^{0.0862}
    \label{eq:joint_lr_bsz_scaliing_LR}
\end{equation}

\begin{table}[h]
    \centering
    \caption{Comparison of scaling laws for optimal batch size and learning rate.}
    \label{tab:batch_size_scaling_laws}
    \begin{tabular}{lcc}
    \toprule
     & Batch size & Learning rate  \\
    \midrule
    \citet{porian2025resolving} & $0.00037\, N^{0.703}$ & $3.7\,N^{-0.36}$ \\
    \citet{li2025steplaw} & $0.58\,D^{0.571}$ & $1.79\,N^{-0.713}\,D^{0.307}$ \\
    \citet{bergsma2025powerlinesscalinglaws} & $0.0306 D^{0.383}$ & -- \\
    This work & $0.000099\, N^{0.145} D^{0.460}$ & $0.371396\, N^{-0.417} D^{0.0862}$ \\
    \bottomrule
    \end{tabular}
\end{table}

\vspace{-2mm}\paragraph{Prediction accuracy on held-out samples.} To assess the predictive power of these scaling laws, we validate them on the held-out 1.7B model. The predicted joint optima closely match the empirical optima observed on the learning rate--batch size grid (\autoref{fig:joint_hp_scaling_preciction_accuracy_grid}), yielding a mean absolute percentage error (MAPE) of 37\% and 22\% for $\eta$ and $b$ respectively. More interestingly, since the regressions are fitted to optima obtained after quadratic smoothing, we repeat the smoothing procedure on the held-out data and compare the predictions against the minima of the fitted quadratics (\autoref{fig:joint_hp_scaling_preciction_accuracy_contour}). The predicted optima closely match these smoothed estimates, with MAPEs of 28\% and 10\% for $\eta$ and $b$, with all predictions falling within the corresponding parametric prediction intervals.

% ------------------------------------------------------------------

\subsection{Scaling Optimal Learning Rate with Arbitrary Batch Size}
\label{sec:scaling_lr_arbitrary_bsz}

% Motivation
\vspace{-2mm}\paragraph{Why arbitrary batch sizes?}
Jointly optimizing the learning rate and batch size is a standard approach and a common setting in recent work~\citep{deepseekv12024,li2025steplaw,bergsma2025powerlinesscalinglaws}. In practice, however, the batch size may be constrained by the available hardware resources, memory capacity, and parallelization strategy, and is therefore often set to \textit{maximize training throughput}. Moreover, the jointly optimal scaling laws derived in the previous section often predict relatively small batch sizes (e.g., only $\sim 1.6$M tokens for a 1.7B-100BT model), which may not fully utilize modern accelerators. In this scenario, when using a different batch size from the jointly optimal one, the scaling laws from \autoref{sec:scaling_jointly_lr_bsz} \textit{no longer prescribe the optimal learning rate}. We therefore study the scaling of optimal LR at arbitrary batch sizes. 
Although scaling the optimal batch size for a given learning rate is not a common scenario in practice, for completeness, we present a symmetric analysis of this setting in \autoref{sec:scaling_bsz_arbitrary_lr}.

\begin{figure}[h]
    \centering
    \begin{subfigure}{0.4\linewidth}
        \centering
        \includegraphics[width=1\linewidth]{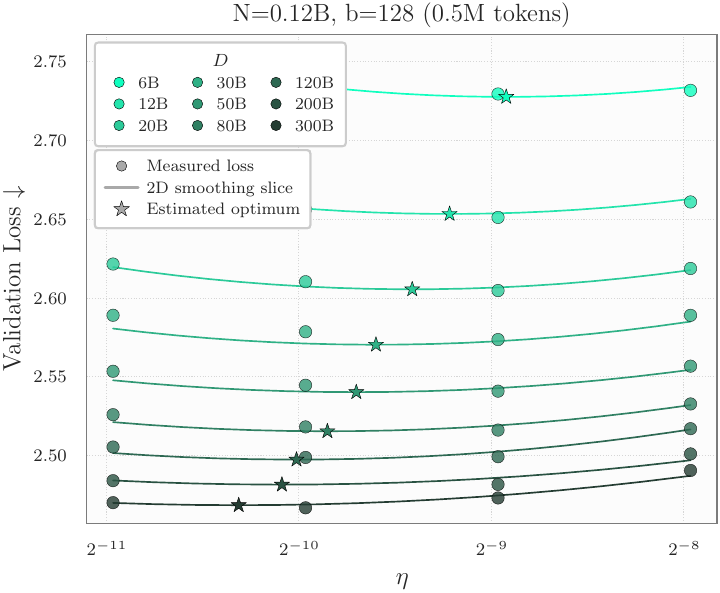}
        \caption{Larger token budgets prefer smaller $\eta^\star$.}
        \label{fig:loss_vs_lr__fix_N}
    \end{subfigure}
    \hspace{1.em}
    \begin{subfigure}{0.4\linewidth}
        \centering
        \includegraphics[width=1\linewidth]{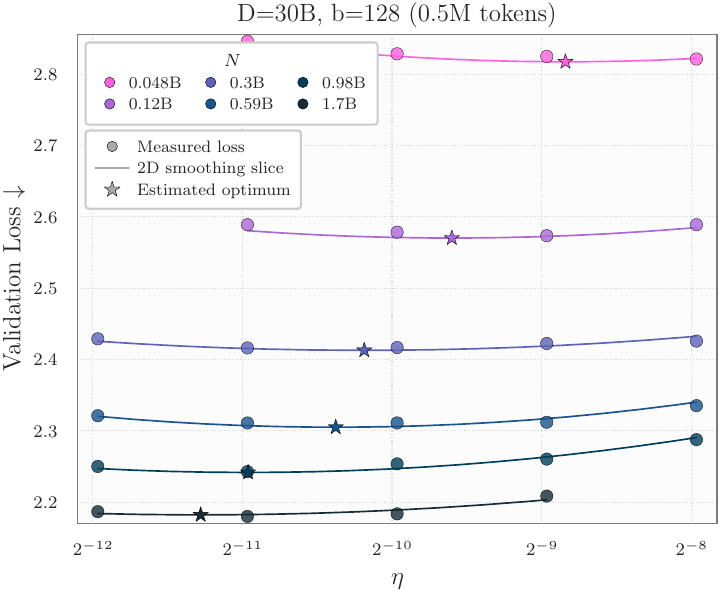}
        \caption{Larger models benefit from smaller $\eta^\star$.}
        \label{fig:loss_vs_lr__fix_D}
    \end{subfigure}
    \caption{Loss as a function of the learning rate $\eta$ at batch size $b=128$, together with the corresponding one-dimensional slices of the fitted two-dimensional quadratic smoothing model. The optimal learning rate decreases as either $D$ or $N$ increases.}
    \label{fig:loss_vs_lr}
\end{figure}

\begin{figure}
    \centering
    \includegraphics[width=1\linewidth]{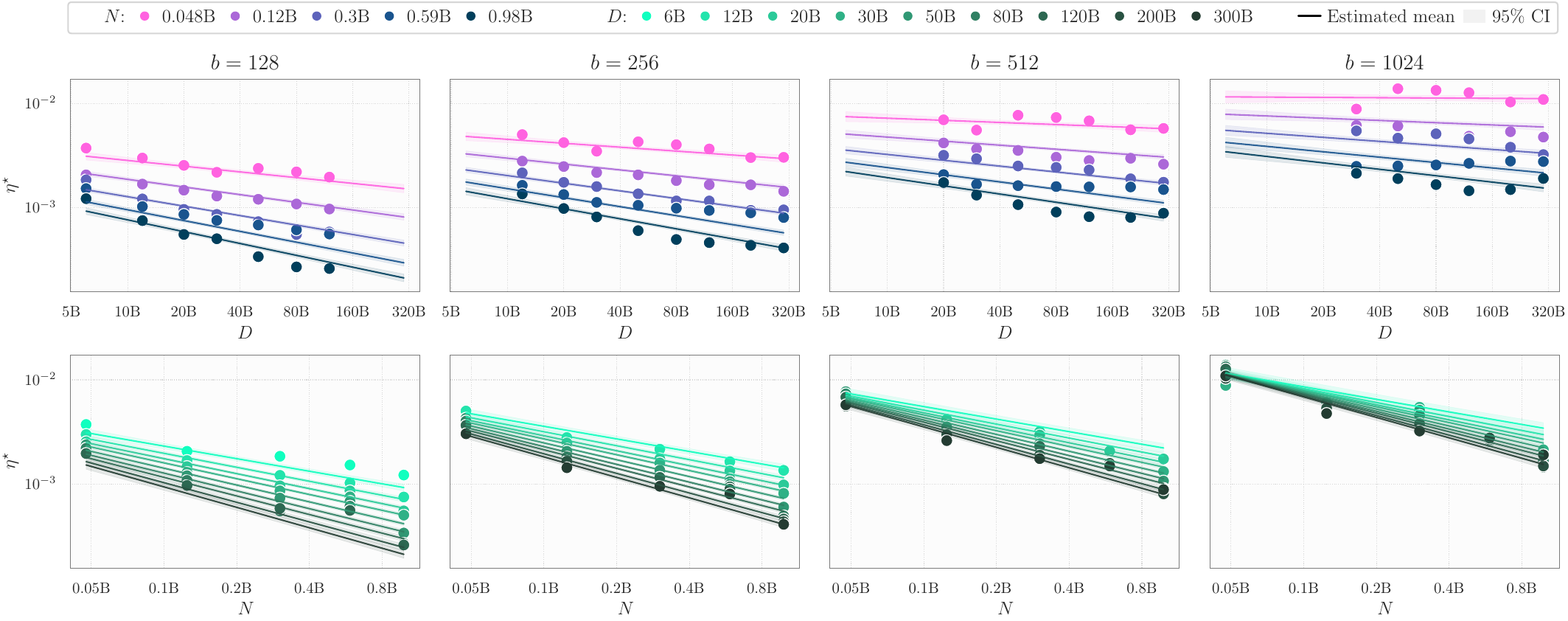}
    \caption{Scaling laws for \textbf{optimal learning rate} at different batch size values, fits from \autoref{eq:scaling_lr_arbitrary_bsz_fit_reduced}.}
    \label{fig:scaling_lr_arbitrary_bsz_fit_reduced_model}
\end{figure}

By constraining $b$, \autoref{eq:quadratic_2d} reduces to a quadratic function of the loss wrt $\eta$, which we depict in \autoref{fig:loss_vs_lr}, and that we use to predict $\eta^\star | b,N,D$ (\autoref{eq:individual_optima}).
We notice that the optimal LR decreases with larger token budgets (\autoref{fig:loss_vs_lr__fix_N}), matching observations from \citet{bjorck2025scalingoptimallrtoken}. For small batch sizes, we see that the optimal LR decreases with model sizes (\autoref{fig:loss_vs_lr__fix_D}), although for larger batch sizes the scaling effect in $D$ weakens substantially (\autoref{fig:scaling_lr_arbitrary_bsz_fit_reduced_model})\footnote{A possible explanation may lie in the interplay between token budget and batch size in determining the total number of optimization steps ($\propto D/b$). At small $b$, increasing $D$ substantially extends the training horizon, potentially favoring smaller and more gradual parameter updates and therefore smaller optimal learning rates. At large $b$ instead, the number of optimization steps remains relatively small even for large $D$, so training may still require aggressive updates.}.

\vspace{-2mm}\paragraph{Model.}
Since trends appear linear after log-log transformations, we once again assume a linear model in log scale with possible interactions between $D$, $N$, $b$:
\begin{equation}
\begin{aligned}
    \log \eta^\star = \,
        & \alpha_0 + \alpha_1 \log b \, + \\
        & \beta_0 \log N + \beta_1 \log b \log N \, + \\
        & \gamma_0 \log D  + \gamma_1 \log b \log D  \, + \\
        & \delta_0 \log N \log D  + \delta_1 \log b \log N \log D  \, + \epsilon.
\label{eq:lr_vs_bND_full_model}
\end{aligned}
\end{equation}
Consistent with previous observations, we find that the optimal learning rate scales strongly with model size ($\hat{\beta}_0\approx-0.54$) and more weakly with data scale ($\hat{\gamma}_0\approx-0.25$). 
Moreover, increasing the batch size shifts the optimal learning rate upward, consistent with the intuition that larger batch sizes permit larger learning rates. The batch size also slightly modifies the dependence on $D$, while leaving the scaling with respect to $N$ essentially unchanged.
We report the estimated coefficients in \autoref{tab:lr_vs_bND_full_model}. After removing non-significant interaction terms, we obtain the following reduced model:
\begin{equation}
    \log \eta^\star = 
    -6.60
    +0.80 \log b
    -0.53 \log N
    -0.24 \log D 
    -0.06 \log N \log D 
    +0.08 \log D  \log b.
\label{eq:scaling_lr_arbitrary_bsz_fit_reduced}
\end{equation}
We show the estimated mean response in \autoref{fig:scaling_lr_arbitrary_bsz_fit_reduced_model}, observing a good fit for small batch sizes, with some deviations at large $b$. 
Compared with the jointly optimal scaling, we find an \emph{opposite trend} with respect to data scaling. When jointly scaling $\eta$ and $b$, larger token budgets favor larger learning rates. In contrast, when optimizing $\eta$ conditionally on $b$, the optimal learning rate decreases with $D$. One possible explanation for this difference comes from the observation that larger token budgets are often associated with larger optimal batch sizes, largely independently of $\eta$, as shown later in \autoref{tab:b_vs_lrND_full_model}. Since larger $b$ are often associated with larger values of $\eta$ for efficient optimization, this may indirectly shift the jointly optimal learning rate upward as $D$ increases.

\begin{figure}
    \begin{subfigure}{\linewidth}
        \centering
        \includegraphics[width=.7\linewidth]{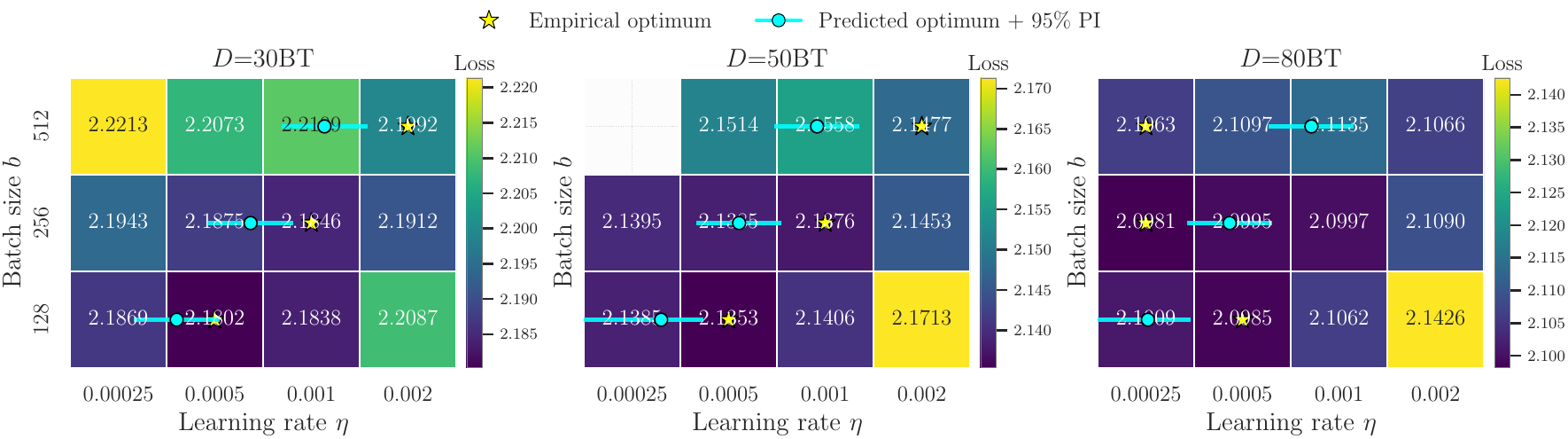}
        \caption{Empirical loss values.}
        \label{fig:lr_vs_bND_prediction_1_7B_grid}
    \end{subfigure}
    \begin{subfigure}{\linewidth}
        \centering
        \includegraphics[width=.7\linewidth]{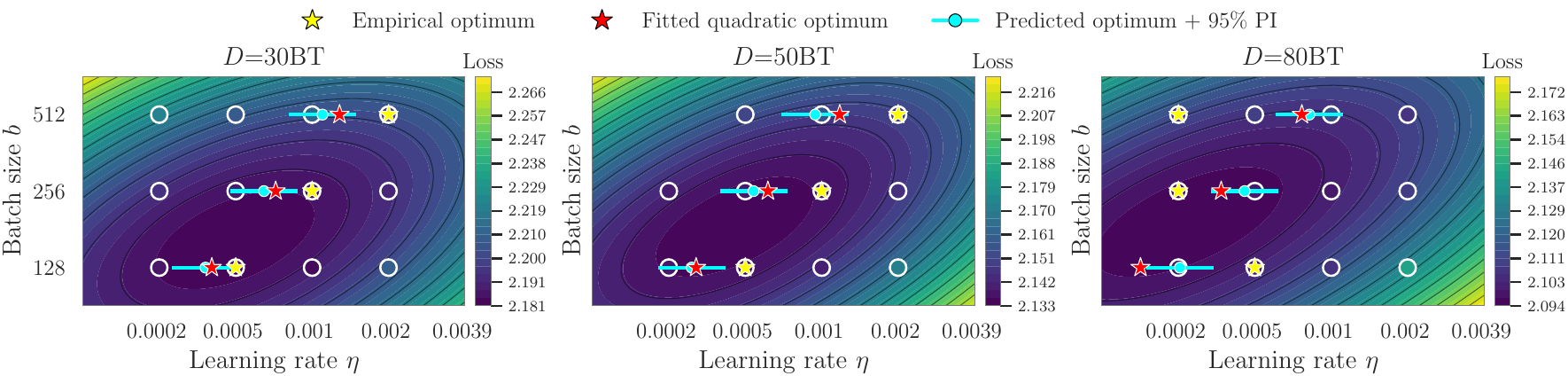}
        \caption{Quadratic approximation of the same loss surface.}
        \label{fig:lr_vs_bND_prediction_1_7B_contour}
    \end{subfigure}
    \caption{%
    \textbf{Prediction accuracy of} \autoref{eq:scaling_lr_arbitrary_bsz_fit_reduced} \textbf{on held-out 1.7B model}. We report empirical optima (\textcolor[HTML]{FF0200}{$\star$}) and minima of the one-dimensional quadratic slices (\textcolor[HTML]{E1D611}{$\star$}). Cyan circles and error bars denote the predicted optima with pointwise parametric $95\%$ prediction intervals. The empirical optima at $D=80$BT deviate from the overall trend, highlighting the noise in these measurements. This also affects the fitted quadratic, which shifts slightly towards smaller $\eta$ values.}
    \label{fig:lr_vs_bND_prediction_1_7B}
\end{figure}

\vspace{-2mm}\paragraph{Prediction accuracy on held-out samples.} 
Finally, we evaluate the model's predictive performance on a held-out 1.7B model trained across a range of batch sizes and token budgets in \autoref{fig:lr_vs_bND_prediction_1_7B}. Relative to the empirical optima, the predicted optimal $\eta$ appear conservative (\autoref{fig:lr_vs_bND_prediction_1_7B_grid}). However, when compared against the smoothed loss surface in \autoref{fig:lr_vs_bND_prediction_1_7B_contour}, we see a much closer agreement, and the pointwise prediction intervals always capture the (estimated) quadratic minima. 

% ------------------------------------------------------------------

\section{LR Annealing Effect on Hyperparameters}
\label{sec:annealing_effect}

% Motivation
Although WSD significantly reduces the computational cost of experiments, learning rate decay runs still amount for an important part of the compute cost (roughly $\mathbf{78.6\%}$ of total compute in our setup).
This cost is further amplified by the need to sweep hyperparameters across model scales, driving up costs and energy consumption. 
A tempting way to reduce this burden is to omit learning rate annealing and instead study the scaling behavior of checkpoints trained with a constant learning rate. However, it is not clear whether (i) LR-decay brings a constant improvement across training configurations ($N,D,\eta,b$), and (ii) if the optimal hyperparameters shift after decay. Recently, \citet{rutte2026scaling_diffusion} argued that learning rate and batch size optima are largely unaffected by annealing, which only shifts the final loss by a constant factor. We revisit this conclusion through a large-scale investigation across our experiments.

\begin{figure}[h]
    \centering
    \includegraphics[width=\linewidth]{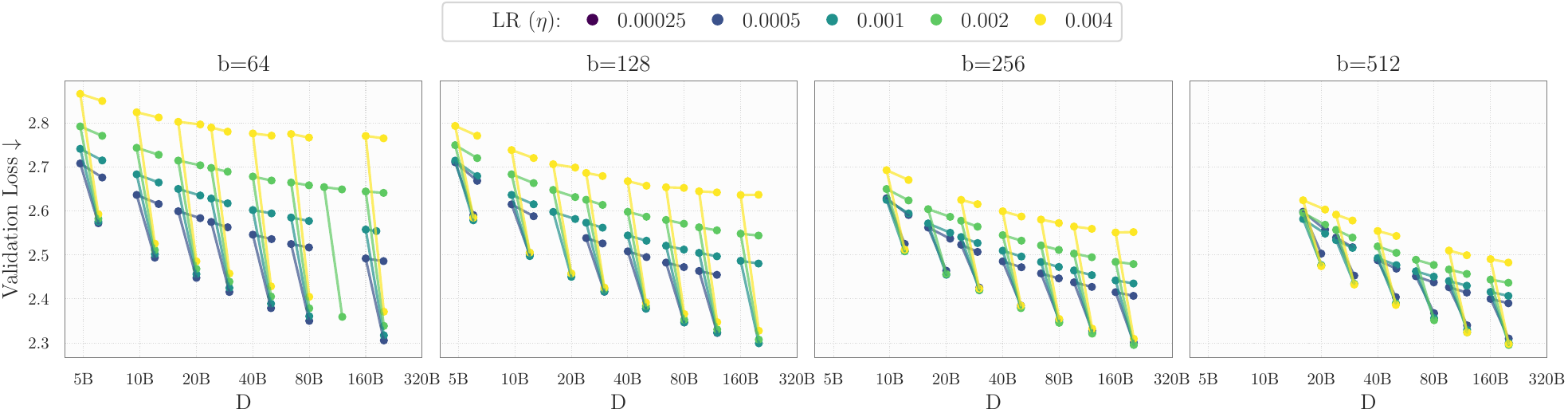}
    \caption{%
    Loss measurements in the two stages of a WSD learning rate schedule across different $D, b, \eta$ values for a $N=0.3$B model. The annealing-induced improvement varies across configurations.
    }
    \label{fig:wsd_loss_segments}
\end{figure}

Before proceeding, we remark that learning rate annealing for a run with a target token budget of $D$ spans the last $20\%$ of tokens\footnote{%
Although this is common practice in the community \citep{bakouch2025smollm3}, different conclusions might hold for a different annealing strategy.}, and ends with a final learning rate of $10^{-5}$. We refer to the loss after decay at a token budget $D$ as $L_{decay}(D)$, the stable-phase loss is instead labeled as $L_{stable}(D)$, and their difference is $\Delta L$. \autoref{fig:wsd_loss_segments} depicts an example of the loss measurements collected for this study.

% ------------------------------------------------------------------

\vspace{-2mm}\paragraph{Loss Improvement from LR Decay depends on $N$, $D$, $\eta$, $b$.}
\begin{figure}[h]
    \centering
    \includegraphics[width=1\linewidth]{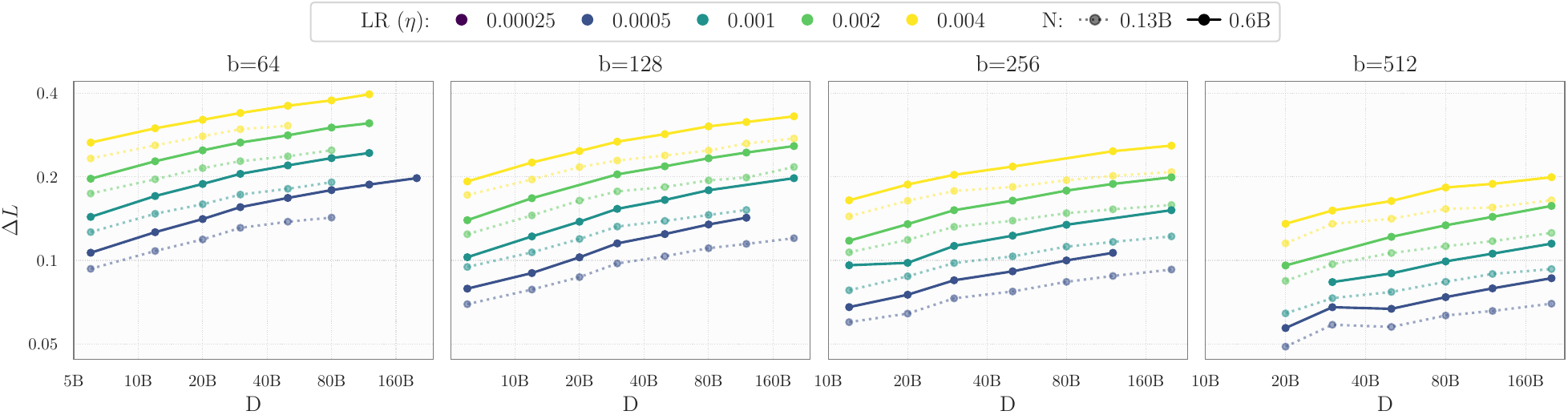}
    \caption{%
    Loss improvement from learning rate annealing for a $0.13$B and $0.6$B model. $\Delta L$ is not constant across $N$, $D$, $\eta$, $b$ values.}
    \label{fig:wsd_delts_loss_vs_D}
\end{figure}
\autoref{fig:wsd_loss_segments} and \autoref{fig:wsd_delts_loss_vs_D} reveal several systematic trends in the loss improvement from decay.
(i) Larger learning rates and smaller batch sizes generally lead to larger loss improvements after decay. This is consistent with the interpretation that larger and smaller batch sizes increase the \textit{noise} in parameter updates, potentially making decay more beneficial. Larger batches also entail fewer parameter updates, potentially reducing the advantage of decay. Beyond this, (ii) the loss reduction is not constant across training horizons $D$, but tends to increase later in training, suggesting that decay may become more beneficial as training progresses. Finally, (iii) this dependence also extends to model size $N$, with larger models exhibiting consistently larger loss improvements from learning rate annealing at the same hyperparameter values.

% ------------------------------------------------------------------
\vspace{-2mm}\paragraph{Optimal Hyperparameters Shift After LR Decay.}

\begin{figure}[h]
    \centering
    \begin{subfigure}{1\linewidth}
        \centering
        \includegraphics[width=\linewidth]{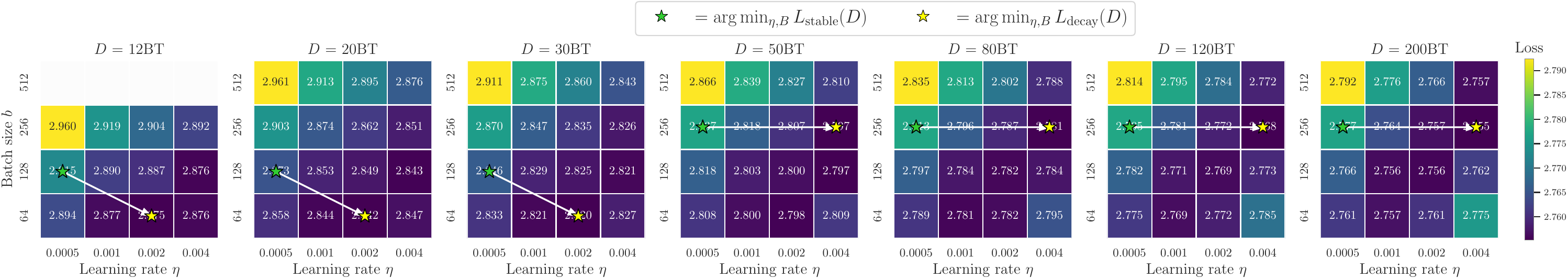}
        \caption{$N=0.05$B}
    \end{subfigure}
    \begin{subfigure}{1\linewidth}
        \centering
        \includegraphics[width=\linewidth]{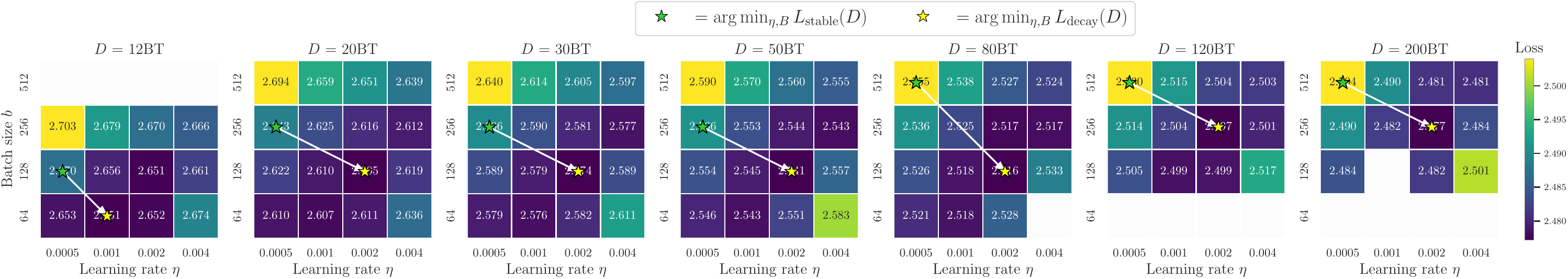}
        \caption{$N=0.13$B}
    \end{subfigure}
    \caption{Joint optimal learning rate and batch size before (\textcolor[HTML]{32CD31}{$\star$}) and after (\textcolor[HTML]{E1D611}{$\star$}) learning rate decay. The heatmap displays the loss \emph{after} learning rate decay ($L_{decay}(D)$) across batch sizes and LR values.}
    \label{fig:hp_shift}
\end{figure}

As the loss improvement from annealing depends strongly on the chosen hyperparameters, unsurprisingly, the optimal hyperparameters also shift between the two training phases (\autoref{fig:hp_shift}).
Specifically, $L_{decay}$ consistently favors larger learning rates, by up to $\sim 4\times$ in our configurations. Smaller batch sizes are also often associated with lower post-decay loss, although this trend is less consistent and has some exceptions. Remarkably, the optimal learning rate for stable runs is consistently the smallest value in our grid, suggesting that smaller learning rates may yield further improvements. These observations persist across different model sizes $N$ and token budgets $D$, and are once again consistent with the idea that using $\eta$ (or smaller $b$) introduces additional optimization noise and favors exploration during training. This additional noise may be beneficial during the stable phase, provided that it is subsequently reduced, for example by decaying the learning rate.

% ------------------------------------------------------------------

\section{Scaling Laws for Performance Prediction}
\label{sec:scaling_loss}

% We now study the scaling behavior of the cross-entropy loss, with three objectives: (i) characterize the scaling properties of our training pipeline, (ii) assess the predictive power of the fitted scaling laws through extrapolation to unseen model and data scales, and (iii) derive power laws fot compute-optimal parameter and data. 

We now study the scaling behavior of the cross-entropy loss, with the aim of characterizing the scaling properties of our training pipeline, and assessing the predictive power of the fitted scaling laws through extrapolation to unseen model and data scales.

To proceed with the analysis, we collect validation loss measurements across several $N$ and $D$, always considering the hyperparameters yielding minimal loss for such $N,D$ configuration\footnote{We use the raw validation loss measurements rather than the smoothed values from \autoref{eq:joint_optima}. The main purpose of smoothing was to obtain reliable estimates of the optimal hyperparameters on the coarse grid to facilitate the derivation of scaling laws, whereas here we prefer the raw measurements to avoid propagating uncertainty from the smoothing fitting process.}.
Unless otherwise specified, we follow a common fitting procedure~\citep{hoffmann2022training, shukor2025scaling, schaipp2026allocatetokensscalinglaws}, minimizing a Huber loss with L-BFGS~\citep{Liu1989LBFGS} with multiple initializations. \autoref{sec:fitting_procedure} provides additional details on the fitting procedure.

\begin{figure}
    \centering
    \begin{subfigure}[t]{0.485\textwidth}
        \centering
        \includegraphics[width=\linewidth]{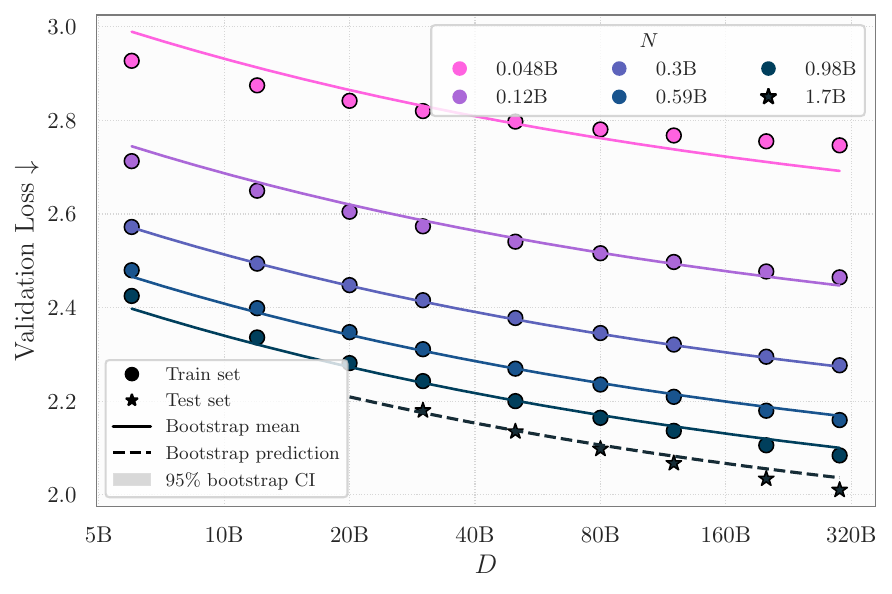}
        \caption{Chinchilla does not model the interaction between $N$ and $D$, leading to poor fit at extreme $N$ and $D$ values.}
        \label{fig:chinchilla}
    \end{subfigure}
    \hfill
    \begin{subfigure}[t]{0.485\textwidth}
        \centering
        \includegraphics[width=\linewidth]{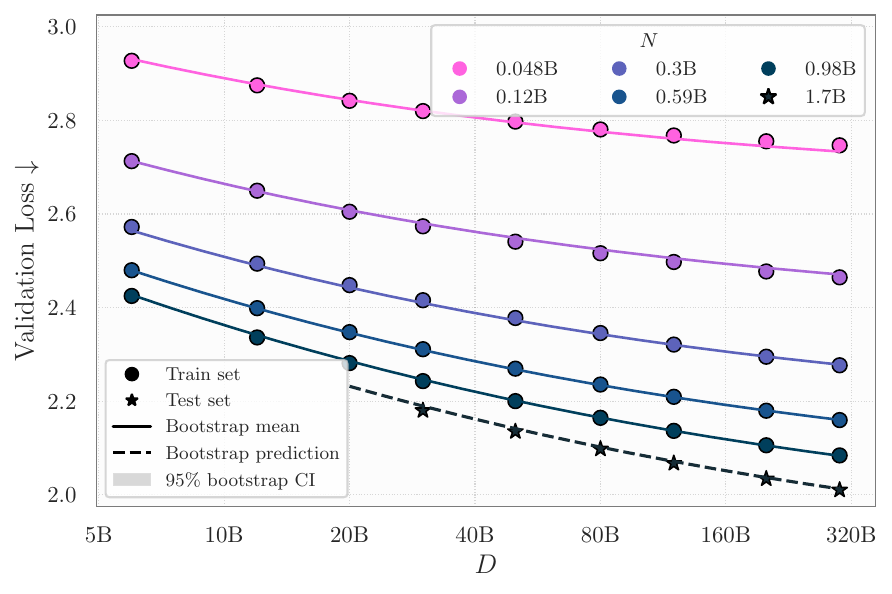}
        \caption{Skaling better captures the interaction between $N$ and $D$, yielding more accurate extrapolation to the $1.7$B model. %larger models can better utilize additional data, while smaller datasets favor lower-capacity models.
        }
        \label{fig:skaling}
    \end{subfigure}
    \caption{Comparison of Chinchilla~\citep{hoffmann2022training} and Skaling~\citep{videau2026skaling} laws.}
    \label{fig:loss_scaling_laws}
\end{figure}

\vspace{-2mm}\paragraph{Chinchilla.}
Following \citet{hoffmann2022training} (see also \cite{rosenfeld2020a}), we first model the loss as the sum of independent power-law contributions from model size $N$ and training data $D$:
\begin{equation}
    L(N, D) = E + A N^{-\alpha} + B D^{-\beta}.
\label{eq:chinchilla}
\end{equation}
Consistent with the observations of \citet{li2025Farseer}, the fitted model in \autoref{fig:chinchilla} provides satisfactory fits for intermediate $N$ and $D$ values, but systematically underperforms for extreme values of either variable, and poorly extrapolates to the 1.7B model.
The residuals also show a clear \textbf{interaction} between $N$ and D: they increase with $D$ for small models but decrease with $D$ for large models, indicating that (i) the benefit of increasing $D$ becomes larger as $N$ increases, and (ii) the benefit of increasing $N$ becomes smaller when $D$ is small. 

\vspace{-2mm}\paragraph{Modeling $N$-$D$ interaction.}
To summarize the previous findings, the effects of model size and training data do not appear fully separable. 
To relax this assumption, we first consider including an additive interaction term, in the form of $L(N, D) = E + A N^{-\alpha} + B D^{-\beta} \textcolor{interaction_color}{+ \, G N^{-\delta}D^{-\gamma}}$. 
Although this form is appealing because it preserves the Chinchilla structure, capturing the observed interaction requires $G<0$, conflicting with the desired monotonicity in N and D~\citep{videau2026skaling}.
We therefore turn our attention to the recently proposed \textit{Skaling} law \citet{videau2026skaling}, which couples model and data capacity through a interaction exponent:
\begin{equation}
    L(N, D) = 
    E + \left( A N^{-\alpha} + B D^{-\beta} \right)^{\textcolor{interaction_color}{k}}.
\label{eq:skaling}
\end{equation}
The extended model (\autoref{fig:skaling}) shows lower Huber loss, and better fits, particulalry at extreme grid observations.
Most importantly, it accurately extrapolates to the held-out 1.7B points, with substantially lower test error (RMSE: 0.0056 vs. 0.0174). We estimate the interaction exponent at $\sim0.41$, close to the values reported by \citet{videau2026skaling}, further supporting an interaction between model size and dataset size beyond the separable Chinchilla law. 
We report the estimated coefficients for both the Chinchilla and Skaling laws in \autoref{tab:loss_scaling_estimated_params}.

\begin{table}
    \centering
    \begin{tabular}{lrrrrrrrr}
    \toprule
    Model
        & $E$
        & $A$
        & $\alpha$
        & $B$
        & $\beta$
        & $k$
        & $\text{Huber}$
        & $\text{RMSE}_{1.7\text{B}}$ \\
    \midrule
    Chinchilla 
        & \begin{tabular}[t]{@{}r@{}}1.418\\[-4pt]\scriptsize $\pm$ 0.25\end{tabular}
        & \begin{tabular}[t]{@{}r@{}}208.5\\[-4pt]\scriptsize $\pm$ 2.1e+02\end{tabular}
        & \begin{tabular}[t]{@{}r@{}}0.2807\\[-4pt]\scriptsize $\pm$ 0.05\end{tabular}
        & \begin{tabular}[t]{@{}r@{}}528.7\\[-4pt]\scriptsize $\pm$ 1.3e+03\end{tabular}
        & \begin{tabular}[t]{@{}r@{}}0.2534\\[-4pt]\scriptsize $\pm$ 0.081\end{tabular}
        & --
        & 1.76e-04
        & 0.0174
        \\
    Skaling 
        & \begin{tabular}[t]{@{}r@{}}1.03\\[-4pt]\scriptsize $\pm$ 0.087\end{tabular}
        & \begin{tabular}[t]{@{}r@{}}2.343e+04\\[-4pt]\scriptsize $\pm$ 1.4e+04\end{tabular}
        & \begin{tabular}[t]{@{}r@{}}0.491\\[-4pt]\scriptsize $\pm$ 0.03\end{tabular}
        & \begin{tabular}[t]{@{}r@{}}4630\\[-4pt]\scriptsize $\pm$ 2.9e+03\end{tabular}
        & \begin{tabular}[t]{@{}r@{}}0.3496\\[-4pt]\scriptsize $\pm$ 0.022\end{tabular}
        & \begin{tabular}[t]{@{}r@{}}0.4105\\[-4pt]\scriptsize $\pm$ 0.033\end{tabular}
        & 3.66e-05
        & 0.0056
        \\
    \bottomrule
    \end{tabular}
    \caption{Estimated scaling law parameters with bootstrap standard errors, together with training Huber loss and held-out RMSE on the 1.7B model.}
    \label{tab:loss_scaling_estimated_params}
\end{table}

\begin{figure}
    \centering
    \includegraphics[width=.55\linewidth]{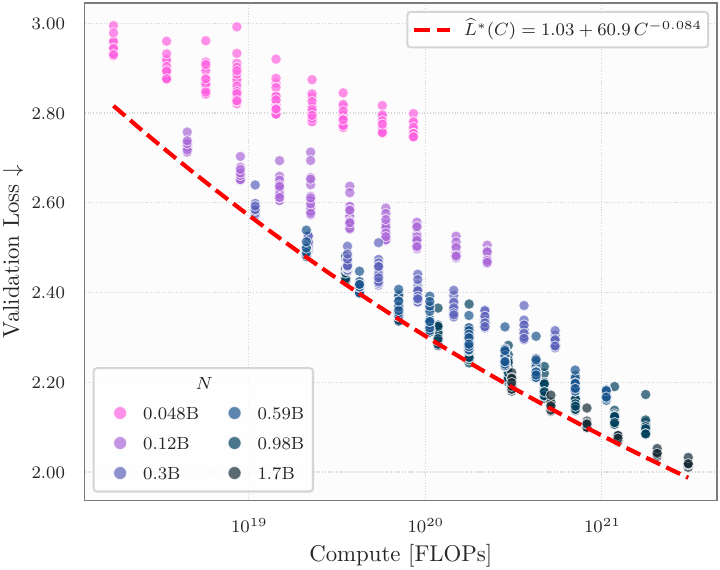}
    \caption{%
    \textbf{Compute-optimal scaling}. Loss observations as a function of compute across all model sizes, data sizes, and hyperparameter configurations, together with the estimated compute-optimal scaling law derived from the parametric functional from  \autoref{eq:skaling}
    }
    \label{fig:loss_vs_C_with_fit}
\end{figure}

\vspace{-2mm}\paragraph{Compute-optimal scaling.}
% Finally, we turn our attention to the evolution of loss as a function of compute. We characterize the scaling of the compute-optimal loss, defined as the minimum loss achievable across $N$ and $D$ configurations under a fixed compute allocation $C$. 
Finally, we turn our attention to characterizing the evolution of compute-optimal loss, defined as the minimum loss achievable across $N$ and $D$ configurations under a compute allocation $C$. 
Given the fitted functional form in \autoref{eq:skaling}, we can derive a closed-form solution for the optimal loss as a function of the $C$\footnote{We refer to \citet{hoffmann2022training} and \citet{videau2026skaling} for the full derivation of the compute-optimal power laws.}. We obtain the following bootstrap mean estimate (see \autoref{fig:loss_vs_C_with_fit}):
\begin{equation}
    L^\star(C) = 1.030 + 60.894\, C^{-0.084}.
    \label{eq:optimal_loss_scaling_law}
\end{equation}

%%%%%%%%%%%%%%%%%%%%%%%%%%%%%%%%%%%%%%%%%%%%%%%%%%%%%%%%%%%%%%%%%%%%%%%%%%%%%%

\section{Downstream Performance}
\label{sec:downstream_evals}

To conclude, we assess the goodness of the pipeline with regards to downstream performance as compute scales. Using \texttt{oellm-eval}\footnote{\url{https://github.com/OpenEuroLLM/oellm-eval}}, we evaluate the models achieving lowest validation loss across $\eta$-$b$ configurations $\forall N,D$. We use the DCLM-CORE evaluation suite \citep{li2024datacomp} and measure downstream performance error. For reference, we additionally evaluate a set of publicly available models under the same setup.
%and compare their performance. 
The set of tasks and evaluation protocol are provided in Appendix \ref{sec:downstream_evals_appendix}, together with the corresponding downstream performance results in Figure \ref{fig:downstream_scaling}. 

\begin{figure}
    \centering
    \includegraphics[width=.7\linewidth]{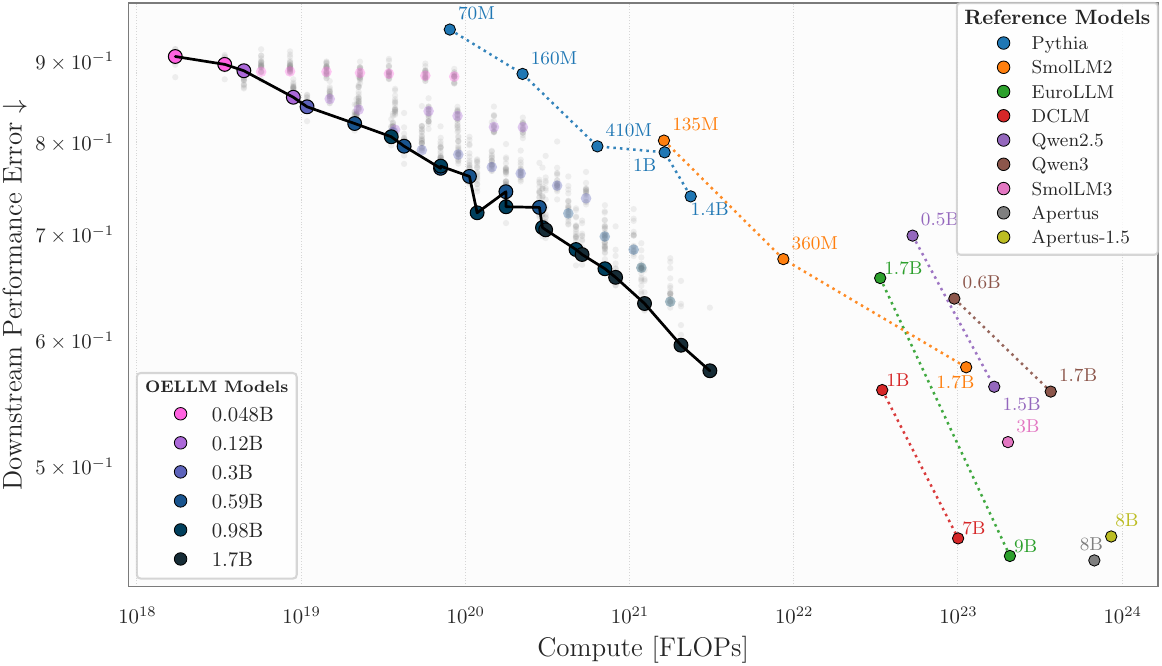}
    \caption{\textbf{Downstream performance error} on the DCLM-CORE evaluation suite as a function of training compute (FLOPs). Faded grey points are all hyperparameter sweep runs, while faded colored points are the per $(N, D)$ lowest training loss configuration that does not sit on the compute-optimal (loss) frontier. OpenEuroLLM models demonstrate consistent performance improvements with increased compute, aligning with the scaling trends of existing open-weights models.}
    \label{fig:downstream_error_scaling}
\end{figure}

\autoref{fig:downstream_error_scaling} shows a consistent decrease in downstream error with increasing compute across all model scales. At the smallest model size, the error plateaus quickly as additional training tokens provide diminishing gains, indicating that model capacity becomes the limiting factor. Larger models continue to benefit from additional compute and follow the overall compute-performance trend observed for other open-weight models. In the compute range where comparison without extrapolation is possible, the OELLM models show better compute efficiency than Pythia and SmolLM2 reference models, achieving lower downstream error at matched compute. This confirms that our proposed training pipeline and hyperparameter optimization efforts generalize consistently across scales and provide a reliable reference point for further model development within OpenEuroLLM. 

%%%%%%%%%%%%%%%%%%%%%%%%%%%%%%%%%%%%%%%%%%%%%%%%%%%%%%%%%%%%%%%%%%%%%%%%%%%%%

\section{Conclusions and Limitations}
\label{sec:conclusions}

This technical report presents an empirical study of scaling laws for large language models. 
We focus in particular on optimal learning rate and batch size scaling, extending the traditional analysis on jointly optimal hyperparameters to also examine their marginal scaling behavior. This is particularly relevant in practical LLM training, where throughput considerations may constrain the batch size away from its jointly optimal value. 
We then investigate how performance scales with respect to model capacity and training tokens, finding that the Chinchilla~\citep{hoffmann2022training} law poorly captures the broad range of training regimes in our experiments, while the Skaling~\citep{videau2026skaling} approach provides better extrapolation to held-out samples through an additional coupling between $N$ and $D$.
%approach that includes multiplicative interaction between N and D provides a better fit, as measured by the error on held-out samples.
Finally, we leverage the employed two-stage Warmup-Stable-Decay LR schedule to investigate open questions about hyperparameter transferability between the two stages. Our analysis reveals a systematic \textit{structure} in how learning rate decay affects the loss, which we believe opens interesting questions about how to model this effect. Successfully \textit{predicting} this evolution could enable more efficient and cheaper scaling law derivation without learning rate decay.
%, as well as to derive compute-optimal $N$-$D$ allocations in a novel way.

There are several limitations to our study. While we find that small-scale scaling laws can extrapolate successfully over roughly an order of magnitude, it remains unclear whether they continue to hold at substantially larger scales \citep{lourie2026smallscaleexperimentsyet}. 
Moreover, further analysis of compute-optimal scaling is necessary to fully characterize these relationships and enable reliable extrapolation to unseen scales. We leave to future work the study of complementary approaches beyond the parametric formulation considered here.
Finally, our work adds to a growing number of studies around optimal hyperparameter scaling, yet the scaling exponents reported across different studies vary substantially, reflecting differences in datasets, architectures, and training configurations. Therefore, transferability across these settings remains limited and warrants further investigation. 

%Moreover, deriving scaling laws is a challenging task, often sensitive to design choices, from data collection to the fitting procedure \citep{li2025misfittingsurveyscalinglaws}, 

%%%%%%%%%%%%%%%%%%%%%%%%%%%%%%%%%%%%%%%%%%%%%%%%%%%%%%%%%%%%%%%%%%%%%%%%%%%%%

\section*{Acknowledgments}
This research was partially supported by the European Commission under the grant No. 101195233 (OpenEuroLLM). JJ acknowledges funding by EU Horizon under grant no. 101214398 (ELLIOT), co-funding from EU under Digital Europe Program under grant no. 101198470 (LLMs4EU) and from EuroHPC Joint Undertaking programme under grant no. 101182737 (MINERVA), funding by the German Federal Ministry of Research, Technology and Space (BMFTR) under grant no. 01IS24085C (OPENHAFM), under the grant 01IS22094B (WestAI -- AI Service Center West), and under the grant 16HPC117K (MINERVA). We acknowledge CSC – IT Center for Science, Finland, for awarding this project access to the LUMI supercomputer, owned by the EuroHPC Joint Undertaking, hosted by CSC (Finland) and the LUMI consortium through the Finnish LUMI Extreme Scale Access programme (OpenEuroLLM Design; project 462000963). We acknowledge the EuroHPC Joint Undertaking for awarding the OpenEuroLLM project access to the LUMI supercomputer, hosted by CSC in Finland through a EuroHPC JU Special Access call. The authors would like to thank the OpenEuroLLM consortium as a whole for fruitful discussions and feedback.

%%%%%%%%%%%%%%%%%%%%%%%%%%%%%%%%%%%%%%%%%%%%%%%%%%%%%%%%%%%%%%%%%%%%%%%%%%%%%

% \clearpage   

\bibliographystyle{plainnat}
\bibliography{bibliography}

%%%%%%%%%%%%%%%%%%%%%%%%%%%%%%%%%%%%%%%%%%%%%%%%%%%%%%%%%%%%%%%%%%%%%%%%%%%%%
\clearpage
\newpage

\appendix

\section{Variability Analysis}
\label{sec:variability_analysis}

\subsection{Estimating Noise}

The primary metric of interest across this work is the expected population loss $\mathbb{E}[L(x,r)]$, where the expectation is taken over samples $x \sim \mathcal{X}$ and over the randomness induced by data ordering and model initialization, which we control via a random seed $r$. In practice, we estimate this quantity by the sample mean loss on a validation set of $n=204800$ samples: $\hat L(r) = \frac{1}{n}\sum_{i=1}^{n} L(x_i,r)$. Such estimator is itself a random variable, and we want to characterize its overall variance.

\begin{figure}[h]
\centering
    \begin{minipage}[b]{0.62\linewidth}
        \centering
        \includegraphics[width=0.97\linewidth]{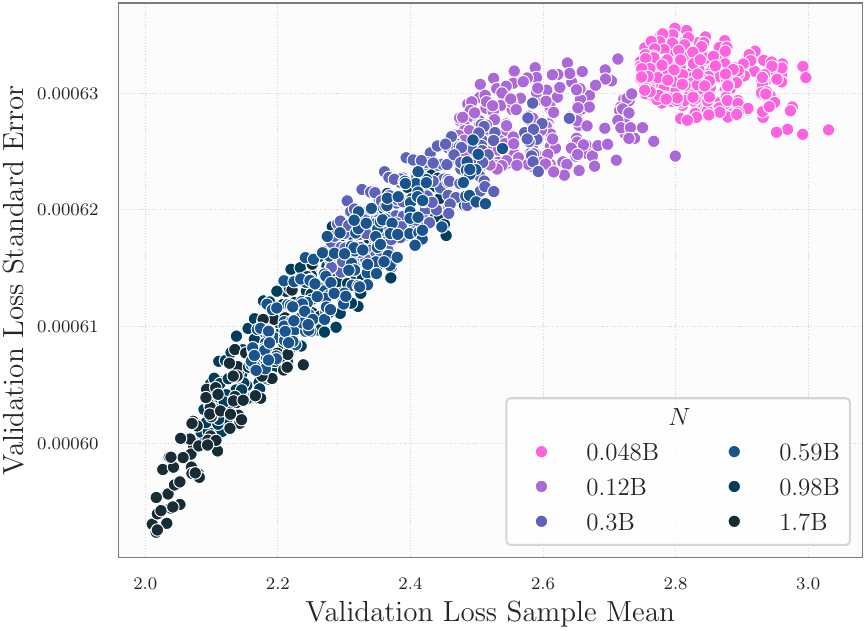}
        \subcaption{Validation loss standard error vs sample mean. Each point represents a training configuration, colored by model size. Larger models exhibit smaller standard errors.}
        \label{fig:finite_sampling_variance_sample_mean_vs_SE}
    \end{minipage}
    \hfill
    \begin{minipage}[b]{0.36\linewidth}
        \centering
        \includegraphics[width=0.9\linewidth]{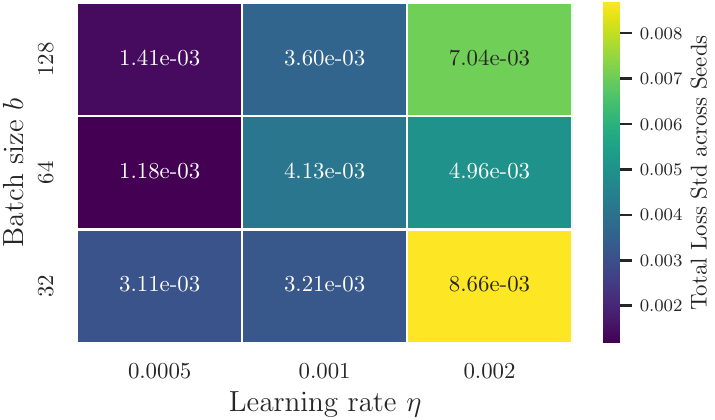}
        \subcaption{Observed validation loss standard deviation across seeds.}
        \label{fig:seed_tot_variance}
        \vspace{0.5em}
        \includegraphics[width=0.9\linewidth]{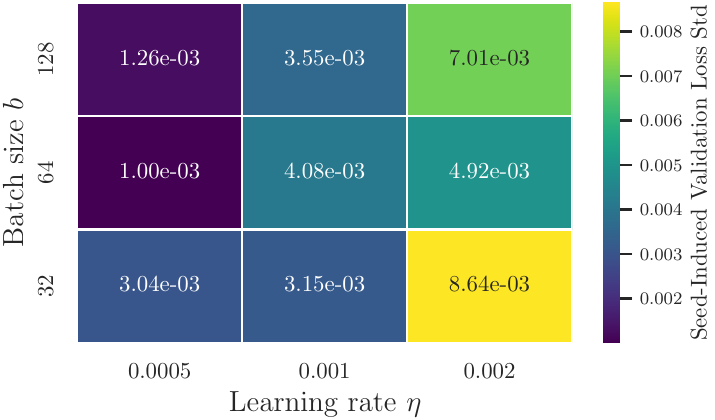}
        \subcaption{Seed-induced validation loss standard deviation.}
        \label{fig:seed_corrected_variance}
    \end{minipage}
    \caption{%
    (\textbf{Left}) Finite-sampling variability.
    (\textbf{Right}) Cross-seed variability for a 50M--6BT model, before and after removing the estimated finite-sampling noise.
    }
\end{figure}

\vspace{-2mm}\paragraph{Finite sampling.}
To estimate the contribution of finite sampling, we fix a random seed, evaluate the loss on each sequence in the validation set, and estimate the standard error of the validation loss mean as
$\widehat{\mathrm{SE}}\!\left[\hat L(r)\right] =
\frac{s_r}{\sqrt{n}}$, where $s_r$ is the standard deviation of the per-sequence losses over the n validation sequences. Similarly to \citep{porian2025resolving}, we observe that the standard error decreases with model size (\autoref{fig:finite_sampling_variance_sample_mean_vs_SE}) and token budget, with more performant models yielding lower-variance validation loss estimates.

\vspace{-2mm}\paragraph{Model initialization and data order.}
In order to estimate the contribution of model initialization and data ordering, we train a $50$M model for $6$B tokens with five different random seeds. The observed variance across runs in \autoref{fig:seed_tot_variance} contains both seed-induced variability and finite-sampling noise, therefore, we subtract the estimated contribution of the latter to isolate the variance introduced by the seed (\autoref{fig:seed_corrected_variance}). For the optimal hyperparameter configuration ($\eta=0.001$, $b=64$) we estimate a total standard deviation of $2.7\cdot10^{-3}$ across runs with different seeds. Of this, $6.3\cdot10^{-4}$ comes from finite sampling of the validation set (which we estimated as in the previous paragraph). Hence, we obtain a seed-induced standard deviation of $2.67\cdot10^{-3}$. We conclude that, at least at this scale and with this validation set size, uncertainty is dominated by seed effects. More interestingly, but not surprisingly, the seed-to-seed variability is up to $3 \times$ larger for higher learning rate or smaller batch sizes, likely because of the additional noise in the optimization process in these configurations.

\subsection{Smoothing Robustness}
\label{sec:smoothing_robustness}

In \autoref{sec:smoothing}, we illustrated how to smooth the loss surface across learning rate and batch size observations, arguing that this procedure yields more robust estimates of optimal hyperparameters. Here, we investigate the robustness of the fitting procedure and of the resulting optimal hyperparameters.

\begin{figure}[h]
    \begin{subfigure}{0.24\linewidth}
        \centering
        \includegraphics[width=\linewidth]{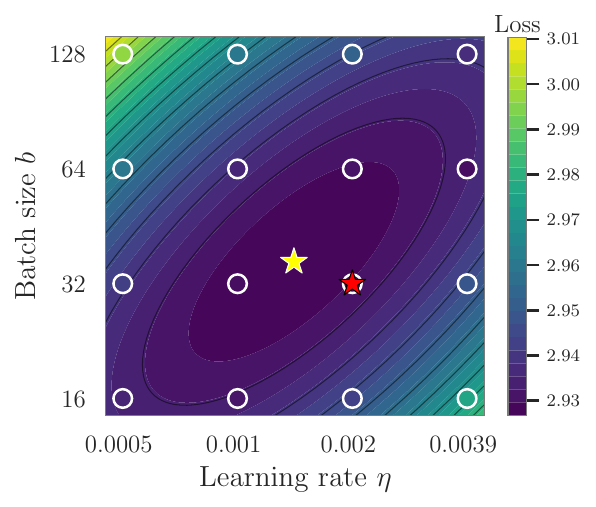}
        \caption*{Seed=1234}
    \end{subfigure}
    \hfill
    \begin{subfigure}{0.24\linewidth}
        \centering
        \includegraphics[width=\linewidth]{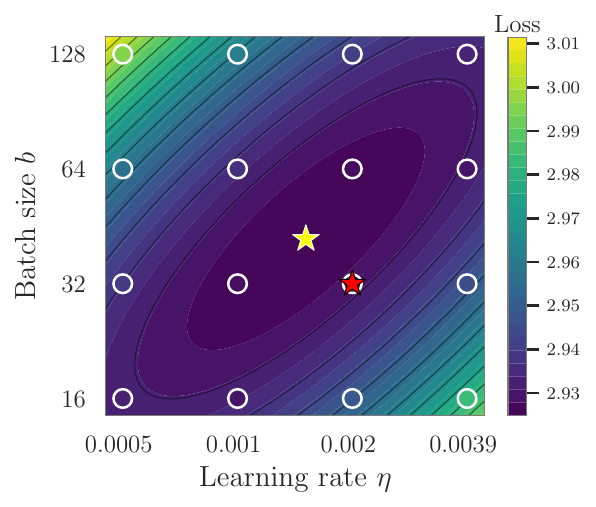}
        \caption*{Seed=1996}
    \end{subfigure}
    \hfill
    \begin{subfigure}{0.24\linewidth}
        \centering
        \includegraphics[width=\linewidth]{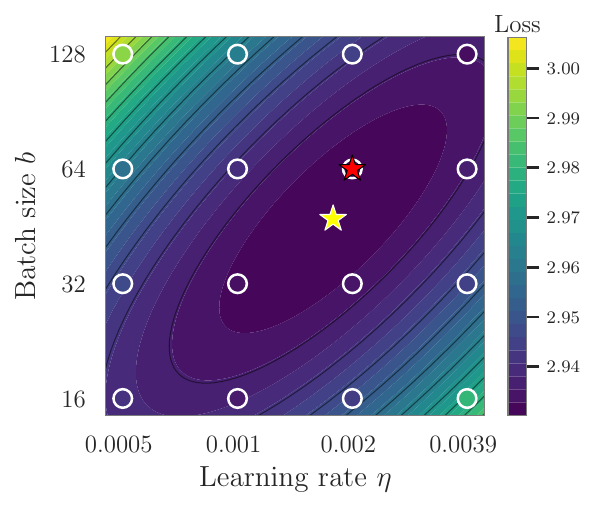}
        \caption*{Seed=2996}
    \end{subfigure}
    \hfill
    \begin{subfigure}{0.24\linewidth}
        \centering
        \includegraphics[width=\linewidth]{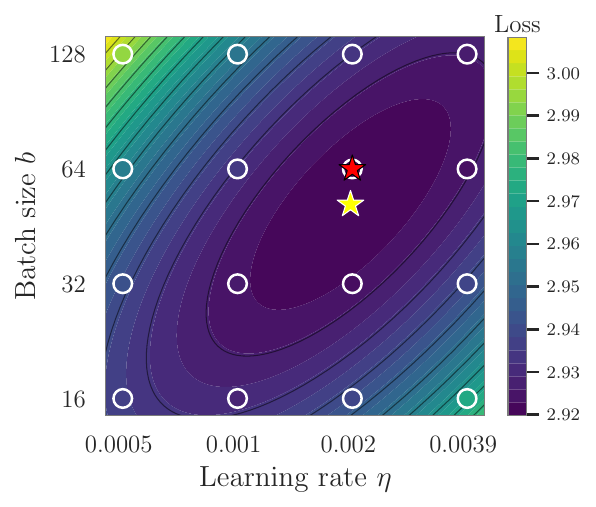}
        \caption*{Seed=3996}
    \end{subfigure}
    \caption{%
    \textbf{Optimal hyperparameters variability}. Contour of the smoothed validation loss surfaces across different rng seeds, with empirical minima (\textcolor[HTML]{FF0200}{$\star$}) and predicted joint optima (\textcolor[HTML]{E1D611}{$\star$}).}
    \label{fig:smoothing_robustness}
\end{figure}

For this analysis, we train a 50M-6BT model across different learning rates and batch size using 4 different random seeds. For each seed, we fit \autoref{eq:quadratic_2d} via OLS, and report the loss contour in \autoref{fig:smoothing_robustness}. This inspection reveals that the \textit{empirical optimal} batch size oscillates between 32 and 64, whereas the optimal learning rate remains stable at the value of 0.002.
We quantify this variability and compare it with that of the \textit{smoothed optima} obtained from the fitted quadratic surface. \autoref{tab:smoothing_robustness} reports the mean and standard deviation of the optimal batch size and learning rate across seeds for both the empirical and smoothed approaches. While both methods identify similar average optima, the smoothed approach reduces the variability of the optimal batch size and introduces a small change in the average predicted learning rate. 

Although limited to a small model, these results suggest that smoothing is not only a robust approach to hyperparameter selection, but may also offer an advantage over direct empirical selection by producing lower-variance hyperparameter estimates. This advantage might be particularly relevant when using a coarse hyperparameter grid, as in our experiments.

\begin{table}[h]
    \centering
    \caption{Empirical and fitted optima across random seeds (mean $\pm$ standard deviation).}
    \label{tab:smoothing_robustness}
    \label{tab:optima}
    \begin{tabular}{lcc}
        \toprule
         & Batch size ($b$) & Learning rate ($\eta$) \\
        \midrule
        Empirical optimum & $48.0 \pm 18.48$ & $0.0020 \pm 0.00000$ \\
        Fitted optimum    & $44.4 \pm 6.56$  & $0.0017 \pm 0.00026$ \\
        \bottomrule
    \end{tabular}
\end{table}

%%%%%%%%%%%%%%%%%%%%%%%%%%%%%%%%%%%%%%%%%%%%%%%%%%%%%%%%%%%%%%%%%%%%%%%%%%%%%

\section{Additional Results}

\subsection{Scaling Optimal Batch Size with Arbitrary Learning Rate}
\label{sec:scaling_bsz_arbitrary_lr}

\begin{figure}[h]
    \centering
        \begin{subfigure}{0.45\linewidth}
            \centering
            \includegraphics[width=1\linewidth]{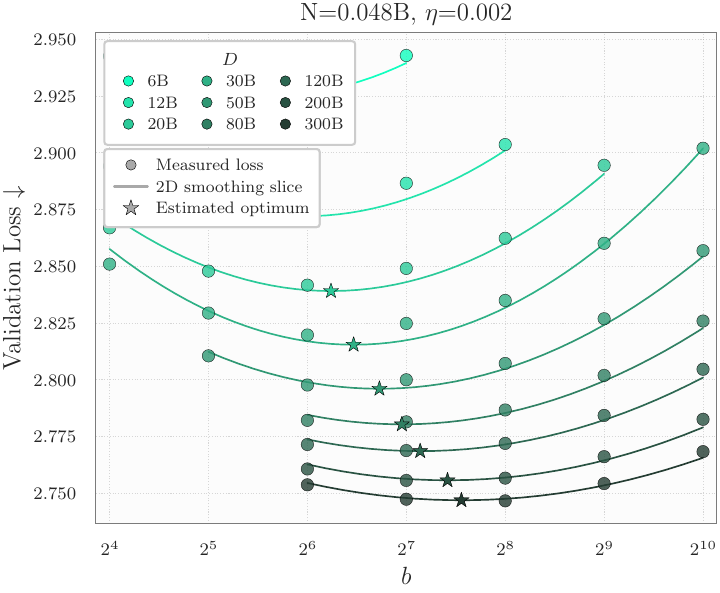}
            \caption{Larger token budgets prefer larger $b^\star$.}
            \label{fig:loss_vs_bsz__fix_N}
        \end{subfigure}
        \hspace{1.5em}
        \begin{subfigure}{0.45\linewidth}
            \centering
            \includegraphics[width=1\linewidth]{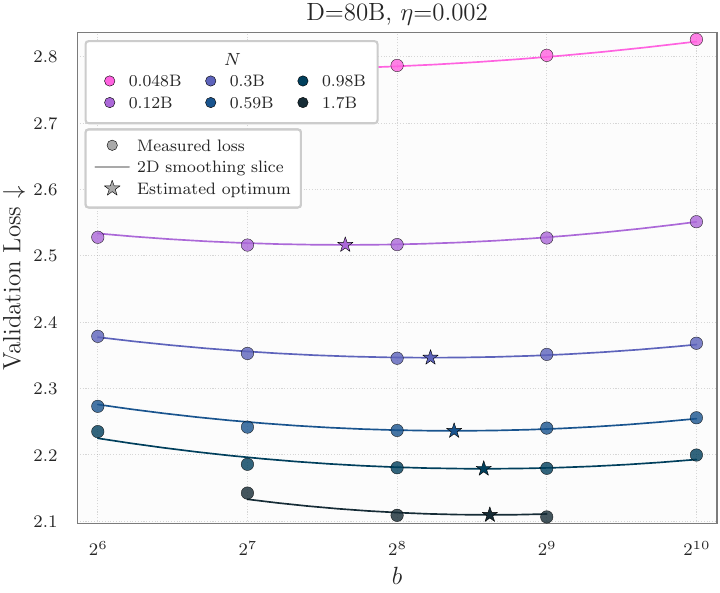}
            \caption{Larger models benefit from larger $b^\star$.}
            \label{fig:loss_vs_bsz__fix_D}
        \end{subfigure}
    \caption{Loss as a function of the batch size $b$ at learning rate $\eta=0.002$, together with the corresponding one-dimensional slices of the two-dimensional quadratic. The optimal batch size increases as either $D$ or $N$ increases.}
    \label{fig:loss_vs_bsz}
\end{figure}

\vspace{-2mm}\paragraph{Model.}
As in \autoref{sec:scaling_lr_arbitrary_bsz}, we plot the loss as a function of the batch size $b$ for different values of $N$, $D$, and learning rate $\eta$, together with the corresponding marginal quadratic, from which we estimate the optimal batch size $b^\star$ (\autoref{eq:individual_optima}).
For a fixed model size $N$, we find the optimal batch size increases with larger token budgets (\autoref{fig:loss_vs_bsz__fix_N}), and for a given $D$, larger models prefer larger batch sizes (\autoref{fig:loss_vs_bsz__fix_D}). Similar patterns persist across different values of $\eta$, $N$ and $D$, and we notice that these trends appear linear in a log-log scale. 
Once again, we hypothesize that the following linear model would adequately represent the data:
\begin{equation}
\begin{aligned}
    \log b^\star = \,
        & \alpha_0 + \alpha_1 \log \eta \, + \\
        & \beta_0 \log N + \beta_1 \log \eta \log N \, + \\
        & \gamma_0 \log D  + \gamma_1 \log \eta \log D  \, + \\
        & \delta_0 \log N \log D  + \delta_1 \log \eta \log N \log D  \, + \epsilon
\label{eq:b_vs_lrND_full_model}
\end{aligned}
\end{equation}
We report the estimated coefficients in \autoref{tab:b_vs_lrND_full_model}.  The first-order $\log N$, $\log D$, and $\log \eta$ terms are all highly statistically significant, confirming that larger models, larger data budgets, and larger learning rates prefer larger batch sizes. In contrast, the $\log D \log \eta$ interaction is weak, suggesting that \textit{the learning rate has little effect on how} $b^\star$ \textit{scales with} $D$. We also find a positive $\log D \log N$ interaction, indicating that the effects of model size and data budget on $b^\star$ reinforce one another. Finally, the three-way interaction term is not statistically significant. 
\begin{table}[h]
    \centering
    \caption{Fitted coefficients and p-values for \autoref{eq:b_vs_lrND_full_model} (n=172, dof=164, $R^2=0.981$).}
    \label{tab:b_vs_lrND_full_model}
    \begin{tabular}{lcc}
    \toprule
    term & coef & pvalue \\
    \midrule
    $\mathrm{const}$ & $5.04673$ & $\approx 0$ \\
    $\log D$ & $0.42089$ & $\approx 0$ \\
    $\log N$ & $0.36163$ & $\approx 0$ \\
    $\log \eta$ & $0.55507$ & $\approx 0$ \\
    $\log D \log N$ & $0.03264$ & $\approx 0$ \\
    $\log D \log \eta$ & $0.01929$ & $0.04132$ \\
    $\log N \log \eta$ & $-0.05628$ & $\approx 0$ \\
    $\log D \log N \log \eta$ & $0.01705$ & $0.05436$ \\
    \bottomrule
    \end{tabular}
\end{table}

\begin{figure}[h]
    \centering
    \includegraphics[width=1\linewidth]{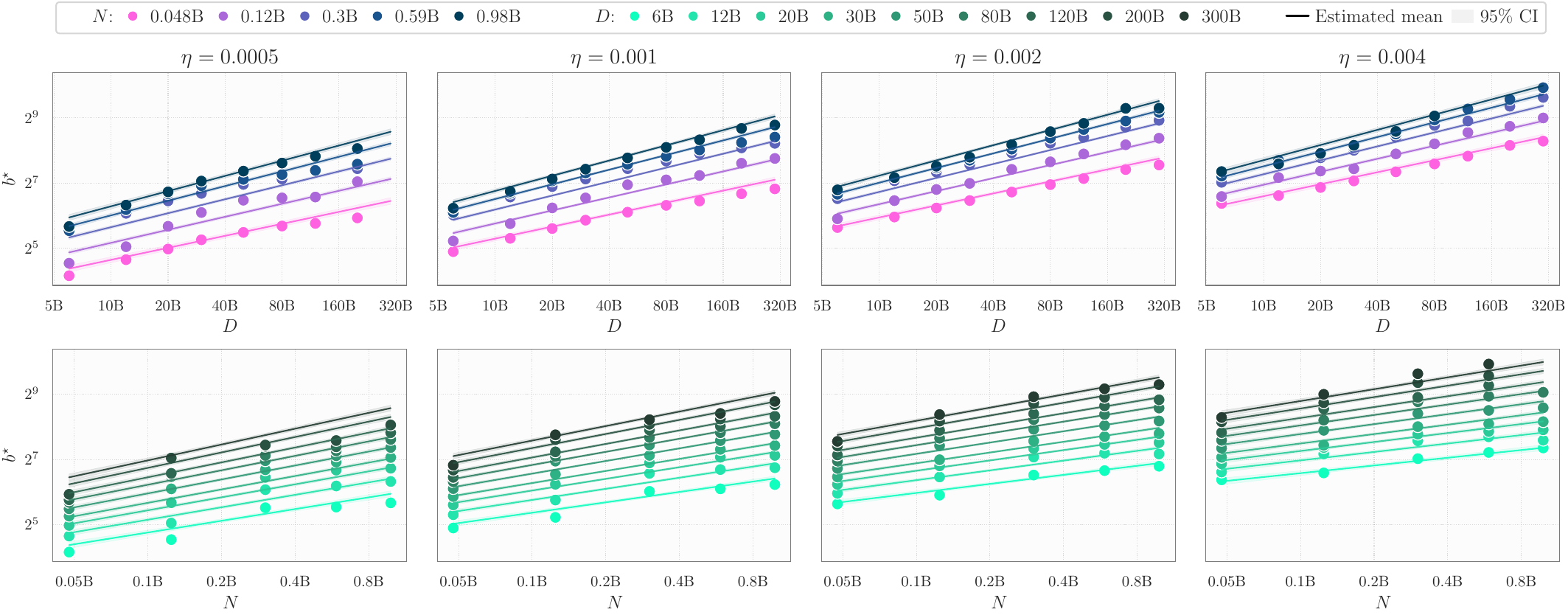}
    \caption{Scaling laws for optimal batch size with arbitrary learning rate values (\autoref{eq:scaling_bsz_arbitrary_lr_fit_reduced}).}
    \label{fig:scaling_bsz_arbitrary_lr_fit_reduced_model}
\end{figure}

After removing non-significant covariates, we obtain the following reduced model (see \autoref{fig:scaling_bsz_arbitrary_lr_fit_reduced_model}):
\begin{equation}
    \log b^\star = 
    5.05
    +0.55 \log \eta
    +0.36 \log N
    +0.42 \log D
    +0.03 \log N) \log D
    -0.06 \log N \log \eta.
    \label{eq:scaling_bsz_arbitrary_lr_fit_reduced}
\end{equation}

\begin{figure}[h]
    \begin{subfigure}{\linewidth}
        \centering
        \includegraphics[width=1\linewidth]{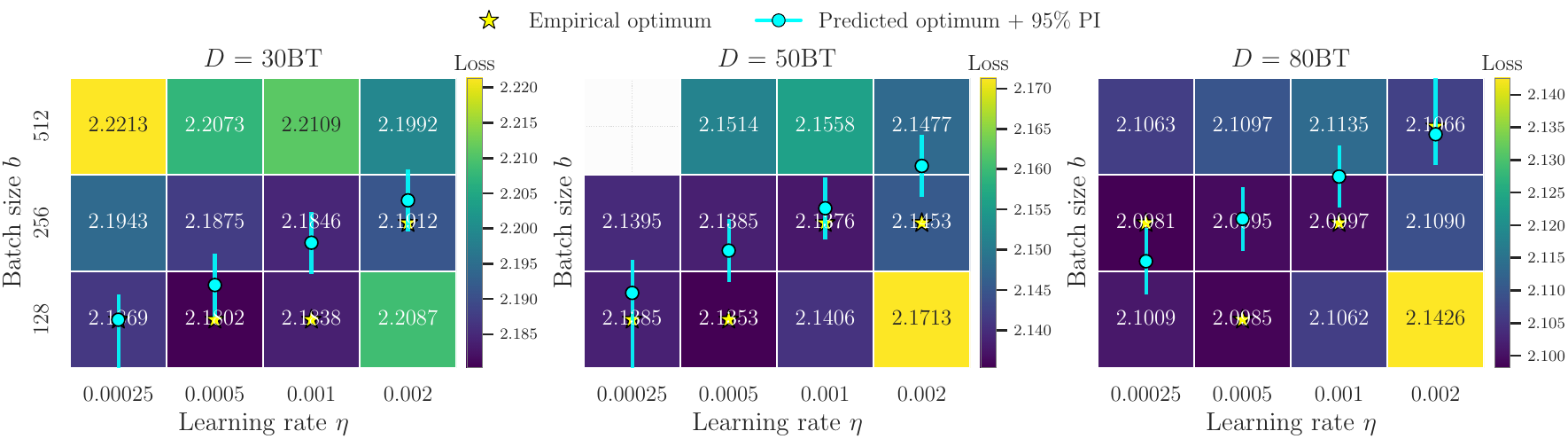}
        \caption{Empirical loss values.}
        \label{fig:b_vs_lrND_prediction_1_7B_grid}
    \end{subfigure}
    \begin{subfigure}{\linewidth}
        \centering
        \includegraphics[width=1\linewidth]{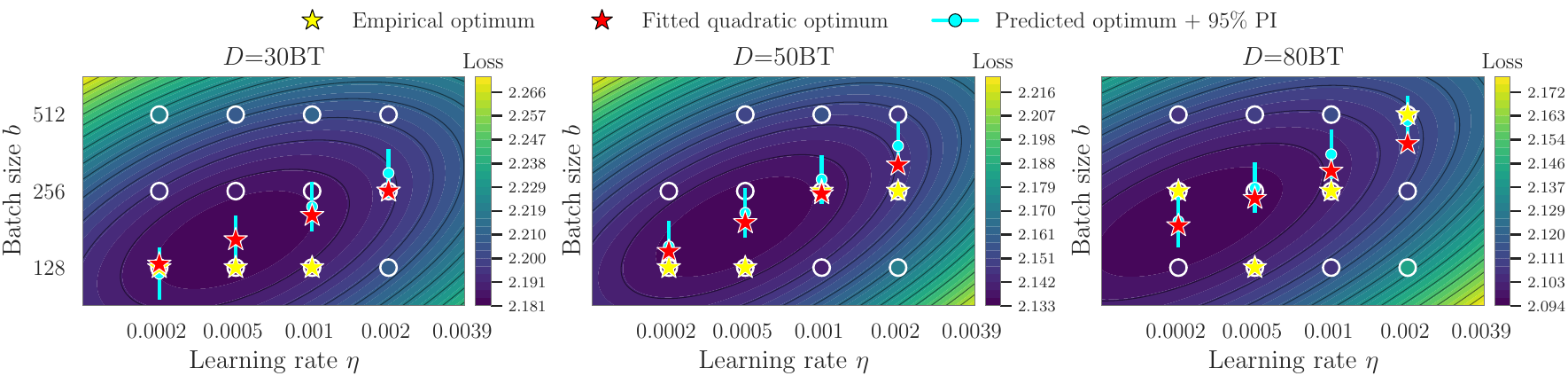}
        \caption{Quadratic approximation of the same loss surface.}
        \label{fig:b_vs_lrND_prediction_1_7B_countour}
    \end{subfigure}
    \caption{%
    \textbf{Prediction accuracy of} \autoref{eq:scaling_bsz_arbitrary_lr_fit_reduced} \textbf{on held-out 1.7B model}. We report empirical optima (\textcolor[HTML]{E1D611}{$\star$}) and minima of the one-dimensional quadratic slices (\textcolor[HTML]{FF0200}{$\star$}). Cyan circles and error bars denote the predicted optima with pointwise parametric $95\%$ confidence intervals at different values of $\eta$ and $D$. Once again, we observe some pronounced noise in the 80BT panel.}
    \label{fig:b_vs_lrND_prediction_1_7B}
\end{figure}

Finally, we test the prediction accuracy on the held-out 1.7B models in \autoref{fig:b_vs_lrND_prediction_1_7B}.

\subsection{Additional Tables}
In \autoref{tab:lr_vs_bND_full_model}, we report the fitted coefficients and p-values for \autoref{eq:lr_vs_bND_full_model}.

\begin{table}[h]
    \centering
    \caption{Fitted coefficients and p-values for \autoref{eq:lr_vs_bND_full_model} (n=186, dof=178, $R^2=0.974$).}
    \begin{tabular}{lcc}
    \toprule
    term & coef & pvalue \\
    \midrule
    $\mathrm{const}$ & $-6.5982501$ & $\approx 0$ \\
    $\log D$ & $-0.2450907$ & $\approx 0$ \\
    $\log N$ & $-0.5410536$ & $\approx 0$ \\
    $\log b$ & $0.7998060$ & $\approx 0$ \\
    $\log D \log N$ & $-0.0635849$ & $\approx 0$ \\
    $\log D \log b$ & $0.0831131$ & $\approx 0$ \\
    $\log N \log b$ & $0.0012659$ & $0.9133354$ \\
    $\log D \log N \log b$ & $0.0154444$ & $0.1171394$ \\
    \bottomrule
    \end{tabular}
    \label{tab:lr_vs_bND_full_model}
\end{table}

%%%%%%%%%%%%%%%%%%%%%%%%%%%%%%%%%%%%%%%%%%%%%%%%%%%%%%%%%%%%%%%%%%%%%%%%%%%%%

\section{Fitting Procedure}
\label{sec:fitting_procedure}

To estimate the parameters of the scaling laws introduced in \autoref{sec:scaling_loss}, we minimize the following log-Huber loss~\citep{besiroglu2024chinchilla}:
\begin{equation}
    \sum_{i=1} \mathcal{H}_{\delta}(\log(L_i) - \log(\hat{L}_i)),
    \qquad 
    \mathcal{H}_{\delta} = 
    \begin{cases}
        \frac{1}{2}x^2 & \quad |x| \leq \delta,\\
        \delta\left(|x|-\frac{1}{2}\delta\right) & \quad |x| > \delta,
    \end{cases}
\end{equation}
where $i$ indexes the training runs, $L_i$ denotes the measured loss and $\hat{L}_i$ is the loss parametrized by either \autoref{eq:chinchilla} or \autoref{eq:skaling}.  We closely follow the methodology in \citet{hoffmann2022training}, optimizing the parameters $E$, $A$, and $B$ in log-space to improve numerical stability, and setting $\delta$ to $10^{-3}$.
% For the Chinchilla model, we use the initialization grid $\log E \in {-1,-0.5,0,0.5,1}$, $\log A,\log B \in {0,5,10,20}$, and $\alpha,\beta \in {0,0.5,1,1.5,2}$. For the Skaling model, we use $\log E \in {-1,-0.5,0,0.5,1}$, $\log A,\log B \in {0,20}$, and $\alpha,\beta,\kappa \in {0,0.5,1,1.5}$.
For the Chinchilla model, we use the following initialization grid:
\begin{equation*}
    \log E \in \{-1,-0.5,0,0.5,1\}, \quad
    \log A,\log B \in \{0,5,10,20\}, \quad
    \alpha,\beta \in \{0,0.5,1,1.5,2\},
\end{equation*}
whereas for the Skaling model, we set:
\begin{equation*}
    \log E \in \{-1,-0.5,0,0.5,1\}, \quad
    \log A,\log B \in \{0,20\}, \quad
    \alpha,\beta,\kappa \in \{0,0.5,1,1.5\}.
\end{equation*}
We repeat the fit for each initialization and select the solution with the lowest final loss. Unlike \citet{videau2026skaling}, we formulate the fitting procedure as an unconstrained optimization problem and solve it using L-BFGS~\citep{Liu1989LBFGS}. To estimate parameter uncertainty, we repeat the fitting procedure, including the initialization sweep, for $1,000$ bootstrap samples, obtained by sampling the training runs with replacement. We report the mean and standard error of the resulting parameter estimates in \autoref{tab:loss_scaling_estimated_params}.

%%%%%%%%%%%%%%%%%%%%%%%%%%%%%%%%%%%%%%%%%%%%%%%%%%%%%%%%%%%%%%%%%%%%%%%%%%%%%

\section{Downstream Evaluation}
\label{sec:downstream_evals_appendix}

The DCLM-CORE benchmark \cite{li2024datacomp} consists of 22 downstream tasks designed to provide a low-variance evaluation signal across different compute budgets. The suite covers world knowledge, commonsense reasoning, language understanding, reading comprehension, and symbolic problem solving. Due to an evaluation error, results for 4 tasks are unavailable at the time of the release; we therefore report performance over the remaining 18 tasks. Table \ref{tab:dclm_core_tasks} summarizes the tasks\footnote{Tasks with missing evaluation scores are marked with *} and few-shot settings, together with their task category and evaluation types. 

\begin{table}[h]
    \centering
    \begin{tabular}{lrll}
    \toprule
    \textbf{Task} & \textbf{Shots} & \textbf{Task Category} & \textbf{Task Type} \\
    \midrule
    AGI Eval LSAT-AR & 3 & symbolic problem solving & multiple choice \\
    ARC Challenge & 10 & world knowledge & multiple choice \\
    ARC Easy & 10 & world knowledge & multiple choice \\
    BIG-bench CS Algorithms* & 10 & symbolic problem solving & language modeling \\
    BIG-bench Dyck Languages & 10 & symbolic problem solving & language modeling \\
    BIG-bench Language Identification & 10 & language understanding & multiple choice \\
    BIG-bench Operators* & 10 & symbolic problem solving & language modeling \\
    BIG-bench QA Wikidata* & 10 & world knowledge & language modeling \\
    BIG-bench Repeat Copy Logic* & 10 & symbolic problem solving & language modeling \\
    BoolQ & 10 & reading comprehension & multiple choice \\
    CommonsenseQA & 10 & commonsense reasoning & multiple choice \\
    COPA & 0 & commonsense reasoning & multiple choice \\
    CoQA & 0 & reading comprehension & language modeling \\
    HellaSwag & 0 & language understanding & multiple choice \\
    HellaSwag & 10 & language understanding & multiple choice \\
    Jeopardy & 10 & world knowledge & language modeling \\
    LAMBADA & 0 & language understanding & language modeling \\
    OpenBookQA & 0 & commonsense reasoning & multiple choice \\
    PIQA & 10 & commonsense reasoning & multiple choice \\
    SQuAD & 10 & reading comprehension & language modeling \\
    Winograd & 0 & language understanding & schema \\
    Winogrande & 0 & language understanding & schema \\
    \bottomrule
    \end{tabular}
    \caption{Downstream evaluation tasks from the DCLM-CORE suite.}
    \label{tab:dclm_core_tasks}
\end{table}

\vspace{-2mm}\paragraph{Model size convention.}
We refer to each model by its publicly reported nominal size. For calculating $C$, we define $N$ following \cite{porian2025resolving} as the number of parameters in the Transformer blocks and output LM head, excluding input embedding parameters for models with untied embeddings. This convention is applied to both OELLM and reference models. 

\begin{figure}
    \centering
    \includegraphics[width=.75\linewidth]{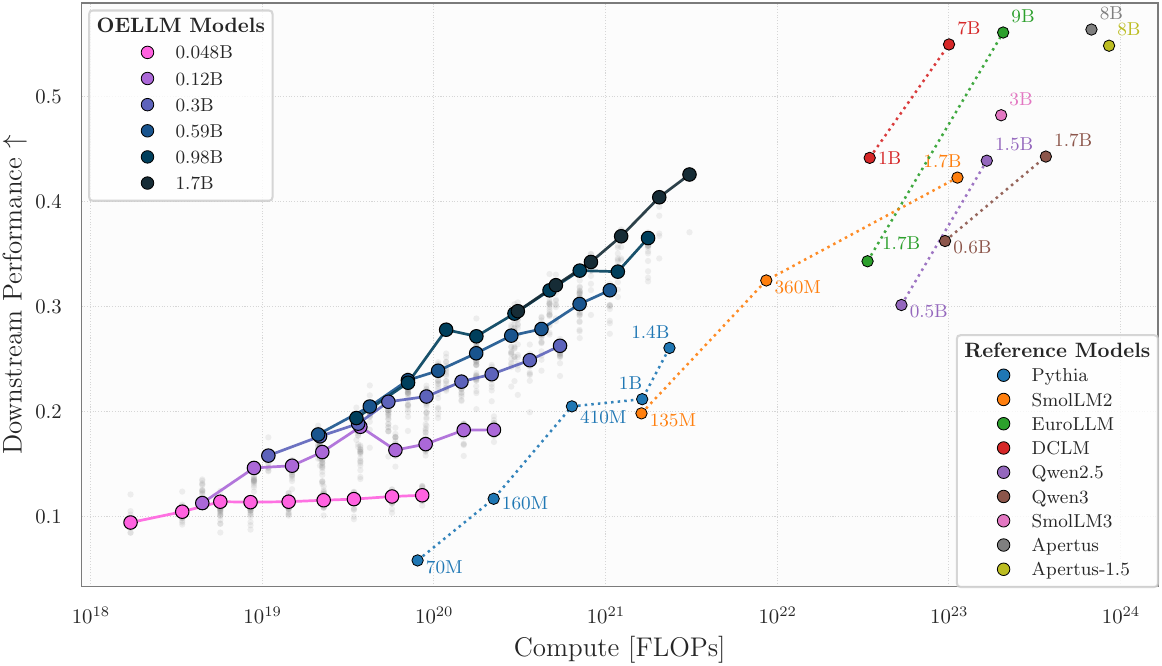}
    \caption{\textbf{Downstream performance} on the DCLM-CORE evaluation suite as a function of training compute (FLOPs). Faded grey points are all hyperparameter sweep runs, while colored points are the per $(N, D)$ lowest training loss configuration for our models. The OpenEuroLLM models exhibit downstream scaling behavior consistent with that of existing open-weight models.}
    \label{fig:downstream_scaling}
\end{figure}

\vspace{-2mm}\paragraph{Metric.}
Following \cite{li2024datacomp}, we use the DCLM-CORE aggregation (\texttt{Core\_v2}). The score of each task is first rescaled in $[0,1]$ and then centered with respect to its random baseline, followed by $[0,1]$ clipping. We clip the centered scores because some evaluation scores fall slightly below the random baseline, resulting in negative centered scores. This maps the random baseline to 0 and a perfect score to 1. Downstream performance is computed as the mean centered score over all available tasks for each model. In Figure \ref{fig:downstream_scaling}, we report downstream error, as $1-\mathrm{DownstreamPerformance}$, such that lower values indicate better performance.

\vspace{-2mm}\paragraph{Reference models.}
To contextualize the scaling behavior of OELLM models, we additionally evaluate publicly available pretrained models spanning a range of model sizes and token budgets. The comparison includes Phytia 70M, 160M, 410M, 1B and 1.4B \citet{biderman2023pythia}; SmolLM2 135M, 360M and 1.7B \citet{allal2025smollm2}; SmolLM3 3B \citet{bakouch2025smollm3}; Qwen2.5 0.5B and 1.5B \cite{qwen25}; Qwen3 0.6B and 1.7B \citet{yang2025qwen3}; EuroLLM 1.7B and 9B \citet{martins2025eurollm, martins2025eurollm9b}; DCLM 1B and 7B \citet{li2024datacomp}; Apertus 1.0 and 1.5 8B \citet{hernandez2026apertus}. All reference models except DCLM are evaluated using the same evaluation setup as the OELLM models; for DCLM, we use the evaluation scores reported in the corresponding Hugging Face model cards~\footnote{%
DCLM-1B model card: \url{https://huggingface.co/TRI-ML/DCLM-1B\#detailed-evaluation}, \\
DCLM-7B model card: \url{https://huggingface.co/apple/DCLM-7B\#evaluation.}
}.
% ~\footnote{
% \href{https://huggingface.co/TRI-ML/DCLM-1B#detailed-evaluation}{DCLM-1B model card};
% \href{https://huggingface.co/apple/DCLM-7B#evaluation}{DCLM-7B model card}}.

For a subset of reference models, one or two task evaluations are unavailable. In these cases, downstream performance is averaged over the available tasks: SQuAD is missing for Qwen2.5-1.5B and Qwen3-1.7B, while both AGI Eval LSAT-AR and SQuAD are missing for Pythia-160M, Pythia-410M, SmolLM2-360M, Qwen2.5-0.5B, Qwen3-0.6B, Apertus-8B, EuroLLM-1.7B, and for EuroLLM-9B.

%%%%%%%%%%%%%%%%%%%%%%%%%%%%%%%%%%%%%%%%%%%%%%%%%%%%%%%%%%%%%%%%%%%%%%%%%%%%%
%%%%%%%%%%%%%%%%%%%%%%%%%%%%%%%%%%%%%%%%%%%%%%%%%%%%%%%%%%%%%%%%%%%%%%%%%%%%%

\end{document}